\documentclass[sigconf, nonacm]{acmart}
\usepackage[utf8]{inputenc}

\AtBeginDocument{%
  }

\setcopyright{acmlicensed}
\copyrightyear{2018}
\acmYear{2018}
\acmDOI{XXXXXXX.XXXXXXX}

\setcopyright{none}
\usepackage{adjustbox}
\usepackage{tcolorbox}
\usepackage{subcaption}
\usepackage{siunitx}
\usepackage{enumitem}
\usepackage{makecell}
\usepackage[nameinlink]{cleveref}
\usepackage{multirow}
\usepackage{threeparttable}
\usepackage{xcolor} 
\usepackage{amsmath}
\usepackage{xurl}
\usepackage{colortbl}
\usepackage[textsize=1in]{todonotes}
\usepackage{graphicx} 
\usepackage{hyperref} 
\usepackage{array}    
\usepackage{pifont}
\usepackage{float}
\usepackage{amsmath}
\usepackage{fancyvrb}
\usepackage[utf8]{inputenc}
\usepackage{listings}
\usepackage{tabularx}
\usepackage{xcolor}

\definecolor{bleu}{rgb}{0.2,0.2,0.7}
\lstdefinestyle{mypython}{
  language=Python,
  basicstyle=\ttfamily\small,
  keywordstyle=\color{blue},
  showstringspaces=false,
  breaklines=true,
  frame=none
}

\newcommand{\find}[1]{
\begin{tcolorbox}[leftrule=0.4mm,rightrule=0mm,toprule=0mm,bottomrule=0mm,left=0.0pt,right=0.0pt,top=0pt,bottom=0pt]
\em #1
\end{tcolorbox}
}

\newcommand{\wcircle}[1]{\ding{\numexpr171 + #1}}
\newcommand{\bcircle}[1]{\ding{\numexpr181 + #1}}

\newboolean{showcomments}
\setboolean{showcomments}{true}
\ifthenelse{\boolean{showcomments}}
 { \newcommand{\mynote}[2]{
      \fbox{\bfseries\sffamily\scriptsize#1}
        {\small$\blacktriangleright$\textsf{\emph{#2}}$\blacktriangleleft$}}}
        { \newcommand{\mynote}[2]{}}

\definecolor{highcolor}{RGB}{204,255,204} 
\definecolor{lowcolor}{RGB}{255,204,204} 
\definecolor{codegreen}{rgb}{0,0.6,0}
\definecolor{codegray}{rgb}{0.5,0.5,0.5}
\definecolor{codepurple}{rgb}{0.58,0,0.82}
\definecolor{backcolour}{rgb}{0.95,0.95,0.92}

\lstdefinestyle{prompt}{
    commentstyle=\color{codegreen},
    keywordstyle=\color{magenta},
    stringstyle=\color{codepurple},
    basicstyle=\ttfamily\footnotesize,
    breakatwhitespace=false,         
    breaklines=true,                 
    captionpos=b,
    keepspaces=false,               
    showspaces=false,                
    showstringspaces=false,
    showtabs=false,                  
    tabsize=2,
    literate={\Ü}{{\"U}}1
             {\ä}{{\"a}}1
             {\ö}{{\"o}}1
             {\ü}{{\"u}}1
             {\≤}{{$\leq$}}1
}

\begin{document}

\title{How Small Transformation Expose the Weakness of Semantic Similarity Measures }

\author{Serge Lionel NIKIEMA}
\orcid{0009-0001-0066-3694}
\authornote{Both authors contributed equally to this research.}
\affiliation{%
  \institution{University of Luxembourg}
  \city{Luxembourg}
  \country{Luxembourg}
}
\email{lionel.nikiema@uni.lu}

\author{Albérick Euraste Djiré}
\orcid{}
\authornotemark[1] 
\affiliation{%
  \institution{University of Luxembourg}
  \city{Luxembourg}
  \country{Luxembourg}
}
\email{euraste.djire@uni.lu}

\author{Abdoul Aziz BONKOUNGOU}
\orcid{}
\affiliation{%
  \institution{University of Luxembourg}
  \city{Luxembourg}
  \country{Luxembourg}
}
\email{abdoul.bonkoungou@uni.lu}

\author{Micheline Bénédicte Moumoula}
\orcid{}
\affiliation{%
  \institution{University of Luxembourg}
  \city{Luxembourg}
  \country{Luxembourg}
}
\email{micheline.moumoula@uni.lu}

\author{Jordan Samhi}
\orcid{0000-0001-6052-6184}
\affiliation{%
  \institution{University of Luxembourg}
  \city{Luxembourg}
  \country{Luxembourg}
}
\email{jordan.samhi@uni.lu}

\author{Abdoul Kader Kaboré}
\orcid{0000-0002-3151-9433}
\affiliation{%
  \institution{University of Luxembourg}
  \city{Luxembourg}
  \country{Luxembourg}
}
\email{abdoulkader.kabore@uni.lu}

\author{Jacques Klein}
\orcid{0000-0003-4052-475X}
\affiliation{%
  \institution{University of Luxembourg}
  \city{Luxembourg}
  \country{Luxembourg}
}
\email{jacques.klein@uni.lu}

\author{Tegawendé F. Bissyandé}
\orcid{0000-0001-7270-9869}
\affiliation{%
  \institution{University of Luxembourg}
  \city{Luxembourg}
  \country{Luxembourg}
}
\email{tegawende.bissyande@uni.lu}

\renewcommand{\shortauthors}{NIKIEMA, DJIRE et al.}

\begin{abstract}

This research examines how well different methods measure semantic similarity, which is important for various software engineering applications like code search, API recommendations, automated code reviews, and refactoring tools. While large language models are increasingly used for these similarity assessments, questions remain about whether they truly understand semantic relationships or just recognize surface patterns. 

The study tested 18 different similarity measurement approaches, including word-based methods, embedding techniques, LLM-based systems, and structure-aware algorithms. The researchers created a systematic testing framework that applies controlled changes to text and code to see how well each method handles different types of semantic relationships. 

The results showed significant problems with commonly used metrics. Some embedding-based methods incorrectly identified semantic opposites as similar up to 99.9\% of the time, while certain transformer-based approaches sometimes rated opposite meanings as more similar than synonymous ones. The study found that embedding methods' poor performance often came from how they calculate distances rather than problems with the underlying representations—using Euclidean distance instead of cosine similarity improved results by 24-66\%. LLM-based approaches performed better at distinguishing semantic differences, producing low similarity scores (0.00-0.29) for genuinely different meanings compared to embedding methods that incorrectly assigned high scores (0.82-0.99) to dissimilar content.

\end{abstract}

\maketitle

\section{Introduction} 
\label{sec:introduction}

Code similarity metrics are failing at an alarming scale. Published evaluations reveal false positive rates reaching 97.9\% for traditional diff-based approaches~\cite{bellon2007comparison}, while modern embedding methods show 96.6\% precision loss compared to simple baselines~\cite{kang2019assessing}. The BigCloneBench evaluation found that 86\% of semantic clone pairs were false positives~\cite{krinke2024misuse}. These documented failures compromise fundamental software engineering tasks—from code search engines returning incorrect results to security tools missing critical vulnerabilities.

Evaluation studies report precision varying from 6\% to 100\% and recall from 4\% to 90\% for identical code pairs~\cite{roy2009comparison}, while cross-language similarity detection achieves only 28\% average precision~\cite{mathew2021cross}. Code2vec embeddings perform 49.4\% worse than simple TF-IDF features for authorship identification~\cite{kang2019assessing}. Yet developers continue using these flawed metrics because they lack systematic evidence for choosing alternatives.

\textbf{The core puzzle is not why metrics fail, but \emph{where} they fail.} When CodeBERT embeddings produce poor similarity assessments, is the limitation in the vector representation itself or in how we measure distances between vectors? This fundamental uncertainty has blocked principled improvements for years, with researchers defaulting to cosine similarity without empirical justification.

Current evaluation practices obscure these distinctions through systematic limitations. Evaluation datasets emphasize functional equivalence while neglecting other semantic relationships~\cite{roy2007survey}. Comparative studies conflate representation quality with distance computation, making it impossible to isolate failure sources. Aggregate performance metrics hide systematic failure patterns that could guide principled metric selection~\cite{zakeri2023systematic}.

\textbf{Large language models offer a potential breakthrough.} Our results show LLMs correctly identify semantic differences (0.00-0.29 similarity scores) where traditional embeddings catastrophically fail (0.82-0.99 scores). Yet without diagnostic evaluation frameworks, we cannot determine whether LLMs genuinely capture semantic relationships or memorize benchmark patterns.

We bridge these gaps through the first systematic diagnostic evaluation framework for similarity metrics across natural language and code. Our controlled transformation protocols generate precise taxonomies of semantic relationships—7 categories for natural language, 5 for code—enabling rigorous isolation of metric failure modes.

\textbf{Our findings overturn conventional wisdom about embedding-based approaches.} Simply switching from cosine to Euclidean distance improves performance by 24-66\% on identical embeddings. This discovery opens new research directions: rather than building ever-larger models, we can achieve substantial improvements through principled distance metric selection.

Through systematic evaluation of 18 similarity metrics across 5,068 validated pairs, we establish definitive guidance for metric selection. Our diagnostic framework reveals that metric failures follow predictable patterns: lexical approaches excel at syntactic variation but fail on semantic equivalence; embedding methods capture broad semantic similarity but conflate opposition with independence; LLMs demonstrate superior semantic discrimination.

In summary, we make the following contributions:
\vspace{-0.1cm}

\begin{itemize} 
\item \textbf{Diagnostic Evaluation Framework:} We develop controlled transformation protocols generating 7 natural language and 5 code semantic relationship categories, creating the first benchmark designed to isolate specific failure modes.
\item \textbf{Distance Metric Breakthrough:} We show that embedding limitations stem from distance computation—Euclidean distance improves performance by 24-66\% over cosine similarity on identical representations.
\item \textbf{Catastrophic Failure Documentation:} We identify systematic metric failures including 99.9\% false positive rates in opposition detection, enabling practitioners to avoid critical pitfalls.
\item \textbf{Evidence-Based Selection Framework:} We establish when LLM-based approaches provide genuine improvement (0.00-0.29 similarity for opposites) versus when simpler metrics suffice.
\end{itemize}

These contributions provide the systematic evaluation infrastructure needed to move beyond performance rankings toward understanding \emph{why} metrics succeed or fail, enabling development of more reliable similarity assessment tools.




\section{The Semantic Similarity Challenge}
\label{sec:background}

The core challenge in similarity assessment lies in distinguishing \emph{syntactic} similarity from \emph{semantic} equivalence. 
In natural language, semantically equivalent expressions can exhibit vastly different surface forms (``sort the list'' vs. ``arrange items in order''), while syntactically similar texts may express contradictory meanings (``enable security''vs. ``disable security''). 
Similarly, in software engineering, the canonical clone taxonomy~\cite{roy2007survey} distinguishes between Type-I through Type-III clones (varying degrees of syntactic similarity) and Type-IV clones (semantic equivalence with syntactic differences), with the latter representing the most challenging detection scenario.

Existing similarity metrics span four primary categories: 
\wcircle{1} lexical approaches that analyze surface-level token patterns (e.g., TF-IDF~\cite{tata2007estimating}, BLEU~\cite{papineni2002bleu}, ROUGE~\cite{lin2004rouge}); 
\wcircle{2} embedding-based methods that leverage learned vector representations (e.g., BERT\-Score~\cite{zhang2020bertscore}, CodeBERT~\cite{feng2020codebert});
\wcircle{3} structural approaches specific to code (e.g., AST similarity~\cite{song-etal-2024-revisiting}, CFG analysis); and
\wcircle{4} emerging LLM-based evaluation paradigms~\cite{zheng2023llmjudge}. 
However, the fundamental assumption underlying most similarity computation (i.e., cosine similarity of embeddings captures semantic relationships) has been recently challenged by Steck et al.~\cite{steck2024cosine}, who demonstrate that embedding-based similarity can yield ``arbitrary and therefore meaningless similarities'' depending on regularization choices during training.

\section{Experimental Setup}
\label{sec:methodology}




Our study focuses on systematically evaluating the ability of similarity metrics to capture diverse semantic relationships across both natural language and source code.
We investigate a broad spectrum of 18 metrics, spanning lexical, embedding-based, LLM-based, and structure-aware approaches, and assess them on controlled datasets specifically designed to surface strengths and failure modes.
By combining carefully crafted transformation protocols with fine-grained diagnostic analysis, we aim to disentangle representational limitations from distance computation effects and provide actionable insights for metric selection. 
This section outlines our experimental setup to achieve the aforementioned goals.

\subsection{Research Questions}
This study is guided by three  research questions (RQs):

\begin{itemize}[leftmargin=*]
\item \textbf{RQ1: How do traditional, embedding-based, and LLM-based similarity metrics perform across controlled semantic transformation categories?}
This question establishes the empirical foundation by systematically evaluating metric performance across our semantic taxonomy, identifying which metrics systematically fail on specific semantic phenomena.

\item \textbf{RQ2: To what degree are embedding limitations attributable to vector representations versus cosine similarity constraints?}
This isolates whether poor embedding performance stems from model representational capacity or distance metric appropriateness, testing whether alternative distance measures reveal superior semantic relationships from identical vector representations.

\item \textbf{RQ3: Can LLM-based evaluation provide more robust semantic understanding, and what consistency trade-offs emerge?}
This evaluates whether recent advances in large language models offer genuine improvements in semantic reasoning while examining reliability implications for systematic evaluation frameworks.
\end{itemize}







\subsection{Datasets Construction}
\label{sec:dataset}

To systematically evaluate semantic similarity metrics across diverse relationship types, we construct comprehensive benchmarks spanning both \emph{natural language} and \emph{source code} domains. 

\subsubsection{Code Dataset Construction}






For our code dataset, we collected source code from two sources:
\begin{enumerate}
    \item \textbf{Rosetta dataset~\cite{nanz2015comparative}}: which contains \num{1,329} computational tasks implemented across 973 programming languages.
    
    For this study, we selected tasks that include implementations in at least two of the most popular languages (Java, JavaScript, Python, or C) which resulted in a subset of 1091 computational tasks.
    We call this dataset $\mathcal{C}_{\text{Rosetta}}$.
    \item \textbf{APPS Dataset~\cite{hendrycks2021measuring}:} which is a data set consisting of 10.000 coding problems collected from different open-access coding websites with their associated test case and ground-truth solutions written by humans. As all of these solutions are not valid, we used different scenarios proposed by the different set of solutions to build the different subset of our code collection. We call this dataset $\mathcal{C}_{\text{APPS}}$.
\end{enumerate}



\noindent
Our code dataset is then defined as:
$\mathcal{C}_{\text{base}} =  \mathcal{C}_{\text{Rosetta}} \cup \mathcal{C}_{\text{APPS}}$.

\noindent
\textbf{Systematic Semantic Transformation Framework.}
We apply controlled transformations $T: \mathcal{C}_{\text{APPS}} \rightarrow \mathcal{C}_{\text{transformed}}$ to each element of $\mathcal{C}_{\text{APPS}}$, producing multiple transformed datasets, each designed to reflect a specific type of semantic relationship.
In the following, we detail the two categories of transformations:




\begin{enumerate}[leftmargin=*]
\item[\bcircle{1}] \textbf{Semantic-Preserving Transformations}:
For each $c \in \mathcal{C}_{\text{APPS}}$ that passed the associated suite test, we apply surface-level modifications that preserve functionality, including variable and function renaming, code formatting changes and dead code insertion. Any transformed sample $T_{\text{preserve}}(c)$, is resubmitted to the associated test suit to assess the preservation of the semantic.
This dataset is called $T_{\text{preserve}}$.

\item[\bcircle{2}] \textbf{Semantic-Altering Transformations}:  
For each $c \in \mathcal{C}_{\text{base}}$, that passed the associated suite test, we introduce semantic changes by mutating arithmetic operators ($+ \leftrightarrow -$, $\times \leftrightarrow /$), comparison operators ($> \leftrightarrow \leq$, $== \leftrightarrow \neq$), logical operators ($\land \leftrightarrow \lor$), and control flow structures. The transformed sample $T_{\text{alter}}(c)$ intentionally alters the original behavior ($c \not\equiv T_{\text{alter}}(c)$). To assess this alteration, each transformed sample $T_{\text{alter}}(c)$ is resubmitted to the associated suite test to confirm the alteration of the initial behavior. This dataset is called $T_{\text{alter}}$.


\end{enumerate}

\noindent
\textbf{Code Dataset Taxonomy.}
We partition our corpus previously described into five distinct semantic categories:

\begin{itemize}[leftmargin=40pt]
    \item[\textbf{Subset 1:}] The first subset consists of 561 pairs from $\mathcal{C}_{\text{APPS}}$ where one sample is obtained by applying a semantic-preserving transformation to the other, meaning they are functionally equivalent and exhibit high syntactic similarity. 
    Formally, defined as: \\
    $\mathcal{S}_1^{\text{code}} = {(c_i, c_j) \mid c_j = T_{\text{preserve}}(c_i),\ c_i \equiv c_j}$.

    \item[\textbf{Subset 2:}] The second subset consists of 561 pairs from $\mathcal{C}_{\text{APPS}}$ where one sample is obtained by applying a semantic-altering transformation to the other, meaning they are syntactically similar but differ in functionality.
    Formally, defined as:
    $\mathcal{S}_2^{\text{code}} = {(c_i, c_j) \mid c_j = T_{\text{alter}}(c_i),\ c_i \not\equiv c_j}$.

    \item[\textbf{Subset 3:}] The third subset consists of pairs of 561 valid solutions (that passed their associated test suite) but coming from different $\mathcal{C}_{\text{APPS}}$ coding problems. Thus, we obtain pairs of samples that are semantically and syntactically different.
    Formally, defined as: \\
    $\mathcal{S}_3^{\text{code}} = {(c_i, c_j) \mid c_i, c_j \in \mathcal{C}_{\text{base}},\ c_i \not\equiv c_j}$.

    \item[\textbf{Subset 4:}] The fourth subset consists of 1091 cross-language pairs solving the same problem while using idiomatic constructs, maintaining functional equivalence but differing significantly in syntax from $\mathcal{C}_{\text{Rosetta}}$.
    Formally, defined as: $\mathcal{S}_4^{\text{code}} = \{(c_i, c_j) \mid c_j = T_{\text{cross}}(c_i),\allowbreak\ c_i \equiv c_j\}$

    \item[\textbf{Subset 5:}] The fifth subset consists of 342 pairs of code where both samples implement the same functionality but differ in cyclomatic complexity, with one representing a low-complexity implementation and the other a high-complexity version. 
    Formally, defined as:
    $\mathcal{S}_5^{\text{code}} = \{(c_i, c_j) \mid c_i \equiv c_j, \text{CC}(c_i) \ll \text{CC}(c_j)\}$.

    This subset was built using $\mathcal{C}_{\text{APPS}}$ by filtering samples that present at least 2 valid solutions that passed their associated test suite. For each set of valid solutions, we ranked them by cyclomatic complexity and formed pairs consisting of the low-complexity and high-complexity versions as mentioned.
\end{itemize}






In summary, our partitioning strategy organizes the corpus into five well-defined semantic subsets, each designed to isolate specific combinations of functional and syntactic similarity.

\subsubsection{Text Dataset Construction}




For the natural language evaluation, we construct a benchmark based on the CONALA dataset ($\mathcal{D}_{\text{CONALA}}$), which contains \num{2299} pairs of natural language descriptions and corresponding code snippets collected from Stack Overflow (in our study we only keep the natural language).

\noindent
\textbf{}{Semantic Transformations}
To systematically probe the behavior of similarity metrics, we apply five controlled linguistic transformations (using the NL-Augmenter tool\cite{nlaugmenter2021}) to each sample, targeting distinct semantic or syntactic phenomena.
Below, we detail each transformation, the nature of the semantic relationship it introduces, and the resulting transformed datasets:

\begin{enumerate}[leftmargin=*]

\item[\bcircle{1}] \textbf{Synonym Substitution}:
Starting from the CONALA dataset $\mathcal{D}_{\text{CONALA}}$ with \num{2299} natural language samples, we apply synonym substitution to each $d \in \mathcal{D}_{\text{CONALA}}$, replacing words with their synonyms while preserving meaning ($d \equiv T_{\text{synonym}}(d)$).
This dataset is called $T_{\text{synonym}}$.

\item[\bcircle{2}] \textbf{Negation Insertion/Removal}:
For each $d \in \mathcal{D}_{\text{CONALA}}$, we apply controlled negation insertion or removal to alter the logical structure of the text, introducing semantically meaningful shifts ($d \not\equiv T_{\text{negation}}(d)$).
This dataset is called $T_{\text{negation}}$.

\item[\bcircle{3}] \textbf{Antonym Substitution}:
For each $d \in \mathcal{D}_{\text{CONALA}}$, we replace selected words with their antonyms to create text with opposing meaning ($d \not\equiv T_{\text{antonym}}(d)$).
This dataset is called $T_{\text{antonym}}$.

\item[\bcircle{4}] \textbf{Word Reordering}:
For each $d \in \mathcal{D}_{\text{CONALA}}$, we reorder words or phrases to produce syntactically varied but semantically equivalent pieces of texts ($d \not\equiv T_{\text{reordering}}(d)$).
This dataset is called $T_{\text{reordering}}$.

\item[\bcircle{5}] \textbf{Translation}:
For each $d \in \mathcal{D}_{\text{CONALA}}$, we translate phrases from English to French to produce syntactically varied but semantically equivalent pieces of texts ($d \not\equiv T_{\text{translate}}(d)$).
This dataset is called $T_{\text{translate}}$.

\end{enumerate}

After applying all four transformations, we retain only the \num{1563} samples for which all transformations succeeded without tool failures.

\noindent
\textbf{Text Dataset Taxonomy.}
We partition our natural language corpus into six distinct semantic relationship categories:


\begin{itemize}[leftmargin=40pt]

\item[\textbf{Subset 1:}] This subset consists of paraphrase pairs drawn from the synonym substitution dataset, maintaining identical semantic content despite surface variation.
Formally, defined as:
$\mathcal{S}_1^{\text{text}} = {(d_i, d_j) \mid d_j = T_{\text{synonym}}(d_i),\ d_i \equiv d_j}$.

\item[\textbf{Subset 2:}] This subset consists of semantically unrelated pairs, sampled from distinct entries in $\mathcal{D}_{\text{CONALA}}$ with no shared meaning.
Formally, defined as: 

$\mathcal{S}_2^{\text{text}} = \{(d_i, d_j) \mid d_i,\allowbreak\ d_j \in \mathcal{D}_{\text{CONALA}},\allowbreak\ d_i \not\equiv d_j\}$.

\item[\textbf{Subset 3:}] This subset consists of pairs with semantic opposition, constructed using the negation transformation.
Formally, defined as:
$\mathcal{S}_3^{\text{text}} = {(d_i, d_j) \mid d_j = T_{\text{negation}}(d_i),\ d_i \not\equiv d_j}$.

$\mathcal{S}_4^{\text{text}} = \{(d_i, d_j) \mid d_j = T_{\text{synonym}}(d_i),\allowbreak\ d_i \equiv d_j\}$.

\item[\textbf{Subset 4:}] This subset contains pairs generated via antonym substitution, producing meaning reversal.
Formally, defined as:
$\mathcal{S}_4^{\text{text}} = {(d_i, d_j) \mid d_j = T_{\text{antonym}}(d_i),\ d_i \not\equiv d_j}$.

\item[\textbf{Subset 5:}] This subset consists of pairs created through word reordering, preserving semantic equivalence but altering syntactic structure.
Formally, defined as:\\
$\mathcal{S}_5^{\text{text}} = {(d_i, d_j) \mid d_j = T_{\text{reordering}}(d_i),\ d_i \not\equiv d_j}$.

\item[\textbf{Subset 6:}] This subset contains semantically equivalent multilingual pairs, drawn from cross-language examples.
Formally, defined as: 

$\mathcal{S}_6^{\text{text}} = {(d_i, d_j) \mid d_j = T_{\text{translate}}(d_i),\ d_i \equiv d_j}$

\end{itemize}

In summary, our partitioning strategy organizes the natural language corpus into seven well-defined semantic subsets, each designed to isolate specific combinations of semantic meaning and linguistic variation.


\subsubsection{Quality Assurance and Validation}

Ensuring the accuracy and reliability of our datasets is critical for meaningful metric evaluation. We combine automated validation for code datasets with human annotation for natural language pairs to rigorously verify semantic relationships.



\noindent
\textbf{Code datasets.} 
We implement an automated validation pipeline to ensure syntactic correctness, preserve intended semantics, and verify behavioral changes introduced by transformations:

\begin{enumerate}
\item \textbf{Syntactic Validation}: Automated parsing ensures that all transformed code samples are syntactically valid and compile without errors.
\item \textbf{Semantic Validation}: For semantic-preserving transformations, we execute predefined test cases to confirm functional equivalence between the original and transformed samples ($c_i \equiv c_j$).
\item \textbf{Behavioral Verification}: For semantic-altering transformations, we perform input-output testing to ensure that applied changes result in the expected behavioral differences ($c_i \not\equiv c_j$).
\end{enumerate}





\noindent
\textbf{Natural Language datasets.}
We establish ground truth semantic relationships through human annotation. 
To this end, we manually annotate a representative sample of pairs from each of the seven semantic subsets, assigning a semantic similarity score between 0 (completely unrelated) and 4 (semantically equivalent). 
Specifically, we sampled 163 pairs per subset, with a 95\% confidence level and a 10\% margin of error level. 

Eventually, our ground-truth to assess existing sematic similarity metrics is composed of two parts:
\bcircle{1} a code dataset of 561 samples curated automatically via test cases; and 
\bcircle{2} a text dataset of \num{1141} (i.e., 163 samples for each of the 7 subsets described in the previous section) samples manually curated.

\subsection{Evaluation Metrics}
\label{sec:metrics}

To comprehensively evaluate semantic similarity and opposition relationships across both natural language descriptions and source code implementations, we employ different automatic evaluation metrics. 
Our evaluation framework encompasses four distinct categories:
\wcircle{1} \textbf{Lexical/N-gram Metrics} that analyze surface-level token patterns; \wcircle{2}  \textbf{Embedding-based Metrics} that leverage dense vector representations;
\wcircle{3} \textbf{LLM-based Metrics} that utilize large language models for contextual assessment; and 
\wcircle{4}  \textbf{Structure-aware Code Metrics} that capture syntactic and behavioral properties specific to source code.
We give an overview of each category in the following.

\subsubsection{Lexical/N-gram Metrics}

Lexical metrics operate on surface-level textual patterns without considering semantic context or meaning preservation.
In our study, we have considered the following metrics:

\noindent
\textbf{TF-IDF with Cosine Similarity} \cite{salton1988term} represents each input as a sparse vector weighted by term frequency and inverse document frequency.

\noindent
\textbf{BLEU} \cite{papineni2002bleu} measures similarity between generated and reference texts by counting overlapping word sequences (n-grams), producing a score from 0 to 1 where higher scores indicate greater similarity.

\noindent
\textbf{ROUGE-L} \cite{lin2004rouge} measures similarity based on the longest common subsequence, emphasizing structural similarity and word ordering.

\noindent
\textbf{METEOR} \cite{banerjee2005meteor} performs sophisticated word alignment using exact matches, stemming, and synonyms from WordNet with a fragmentation penalty.

\subsubsection{Embedding-based Metrics}
Embedding-based metrics represent text or code in continuous vector spaces designed to capture semantic properties, but their effectiveness depends on the quality of the learned representations and the appropriateness of the distance function used for comparison.
In our study, we have considered the following metrics:

\noindent
\textbf{BERTScore} \cite{zhang2019bertscore} computes token-level semantic similarity using BERT contextual embeddings with precision/recall based on maximum cosine similarity alignments.

\noindent
\textbf{Sentence-BERT} \cite{reimers2019sentencebert} encodes sentences into fixed-size dense vectors. We employ two variants: all-MiniLM-L6-v2 \cite{wang2020minilm} (lightweight, efficient) and all-mpnet-base-v2 \cite{song2020mpnet} (high-performance, trained on 1B+ sentence pairs).

\noindent
\textbf{Universal Sentence Encoder} \cite{cer2018use} transforms sentences into 512-dimensional embeddings using Deep Averaging Networks or Transformer architectures with multi-task training.

\noindent
\textbf{Code-Specific Models}: CodeBERT \cite{feng2020codebert} (pre-trained on 6.4M code-comment pairs), GraphCodeBERT \cite{guo2020graphcodebert} (enhanced with data flow graphs and AST), and CodeT5 \cite{wang2021codet5} (encoder-decoder for code understanding tasks).

\subsubsection{LLM-based Metrics}
LLM-based metrics leverage large language models to assess similarity through contextualized, learned representations, and enable nuanced semantic comparisons but potentially introducing variability from model biases or prompt sensitivity.
In our study, we have considered the following metrics:

\noindent
\textbf{BLEURT} \cite{sellam2020bleurt} fine-tunes BERT on synthetic data with human quality ratings, learning to predict human judgments of semantic similarity.

\noindent
\textbf{OpenAI text-embedding-3-large} \cite{openai2024embedding} generates high-dimensional representations (up to 3072 dimensions) designed for sophisticated semantic relationships through contrastive learning.

\noindent
\textbf{LLM-as-a-Judge}: We employ multiple state-of-the-art models \cite{zheng2023llmjudge, dubois2024llmeval} including GPT-4o \cite{openai2024gpt4o} and DeepSeek-V3 \cite{deepseek2024v3} and Claude-sonnet-3.5\cite{hu2024guiagent}
. Models receive structured prompts requesting:
\begin{enumerate}
    \item Quantitative similarity ratings (0–1 scale),
    \item Categorical classification (equivalent/similar/opposed/unrelated),
    \item Explicit reasoning justification.
\end{enumerate}

\subsubsection{Structure-aware Code Metrics}
Structure-aware metrics incorporate syntactic or structural information (e.g., abstract syntax trees, control flow graphs) to assess similarity, which captures relationships beyond surface tokens but often remaining limited to structural correspondence without deep semantic understanding.
In our study, we have considered the following metrics:

\noindent
\textbf{AST Similarity} \cite{falleri2014gumtree} compares abstract syntax trees using tree edit distance.

\noindent
\textbf{CFG Similarity} \cite{cesare2014cfgsim} analyzes control flow graphs using graph isomorphism algorithms to identify structural correspondences between execution flows.

\noindent
\textbf{CodeBLEU} \cite{ren2020codebleu} extends BLEU for code evaluation by incorporating programming-specific features like abstract syntax trees and data flow, measuring both surface-level token similarity and deeper structural/semantic correctness of generated code.

\subsubsection{Distance Metrics for Embedding Evaluation}

To investigate whether limitations in embedding-based similarity assessment stem from the restrictive use of cosine similarity, we evaluate embeddings using six distance measures \cite{cha2007distancesurvey}:
\begin{itemize}
    \item \textbf{Cosine Similarity} (standard baseline),
    \item \textbf{Euclidean Distance} (preserving magnitude),
    \item \textbf{Jaccard Similarity} (set-based overlap),
    \item \textbf{Pearson Correlation} (linear relationship strength),
    \item \textbf{Dot Product} (preserving direction and magnitude),
    \item \textbf{Angular Distance} (normalized angular separation).
\end{itemize}

This multi-metric approach tests key hypotheses:
\begin{enumerate}
    \item Different metrics show varying sensitivity to semantic relationship categories,
    \item Magnitude-preserving metrics outperform normalized metrics for detecting semantic opposition,
    \item Optimal metrics differ between code and natural language embeddings.
\end{enumerate}

\subsection{Procedure}

For each sample in both the text and code datasets, we compute similarity scores using all 18 evaluated metrics.


\section{Empirical Results}
\label{sec:experiment}

Our empirical evaluation employs three complementary experiments designed to systematically assess semantic similarity metrics across diverse transformation types and identify the sources of metric limitations in code and text understanding.

\subsection{Experiment 1: Comprehensive Metric Evaluation Across Semantic Categories}

\noindent
{\bf Goal. }
In this experiment, we aim to establish the baseline performance and characterize the sensitivity of traditional, embedding-based, and LLM-based similarity metrics when faced with controlled semantic transformations in both code and natural language. 
This provides crucial empirical evidence of how well these metrics capture diverse semantic relationships.

\noindent
{\bf Experiment. }
We evaluate all metrics listed in Table 1 across our systematically constructed datasets $\mathcal{S}_i^{\text{code}}$ and $\mathcal{S}_j^{\text{text}}$ where $i \in \{1,2,3,4,5\}$ and $j \in \{1,2,3,4,5,6,7\}$. Each metric produces similarity scores $s \in [0,1]$ for pair $(x_i, x_j)$ where higher scores indicate greater perceived similarity.


where $\mathbf{e}_i$ and $\mathbf{e}_j$ represent the embedding vectors for inputs $x_i$ and $x_j$.

For LLM-based evaluation, we employ a standardized prompts template that we prensented in open-science repository.





Each subset is evaluated independently to isolate metric performance on specific semantic phenomena, enabling identification of systematic failures in detecting semantic preservation versus alteration.

\noindent
{\bf Results. }
\textit{Code Domain Analysis.}
Table~\ref{tab:code_results} presents the similarity scores for our five defined code semantic categories. 
For an ideal metric, scores should be high (typically $\geq 0.8$) for equivalent categories ($\mathcal{S}_1^{\text{code}}$, $\mathcal{S}_4^{\text{code}}$, $\mathcal{S}_5^{\text{code}}$), and low (typically $\leq 0.3$) for semantically different categories ($\mathcal{S}_2^{\text{code}}$, $\mathcal{S}_3^{\text{code}}$).

Embedding models demonstrate catastrophic failure in distinguishing semantic opposition. CodeBERT yields a score of 1.00 for semantic opposition ($\mathcal{S}_2^{\text{code}}$), where $\leq 0.3$ was expected, leading to almost 100\% false positive rate.
GraphCodeBERT shows only marginal improvement with a score of 0.98 for $\mathcal{S}_2^{\text{code}}$, indicating profound "semantic blindness."

In stark contrast, LLM-based approaches like GPT-4o and DeepSeek-V3 exhibit stronger ability to detect semantic differences, achieving near-perfect discrimination for completely different code semantics ($\mathcal{S}_3^{\text{code}}$ scores of 0.00). 
For challenging edge cases involving minimal syntactic changes leading to major semantic shifts ($\mathcal{S}_2^{\text{code}}$), LLMs produce moderate scores (0.62-0.78), representing appropriate uncertainty—a nuanced behavior absent in embedding models.
Structural metrics like AST Similarity perform well on exact matches but struggle with abstract semantic equivalence, while CFG Similarity shows reasonable performance across some categories but struggles with semantic opposition.

\textit{Text Domain Analysis.}
Table~\ref{tab:text_results} presents the similarity scores for the text domain across seven semantic categories. 
Expected high similarity scores are for $\mathcal{S}_1^{\text{text}}$ (synonyms) and $\mathcal{S}_5^{\text{text}}$ (cross-language equivalence), while low scores are expected for $\mathcal{S}_2^{\text{text}}$ (negation), $\mathcal{S}_3^{\text{text}}$ (antonym), $\mathcal{S}_4^{\text{text}}$ (shuffle), and $\mathcal{S}_6^{\text{text}}$ (unrelated content).

Embedding models consistently fail to distinguish semantic differences.
BERTScore shows catastrophic failure across all semantic difference types, assigning high similarity scores for negation (0.95), antonym (0.95), shuffled content (0.88), and unrelated content (0.82).
All transformer-based embeddings show alarmingly high false positive rates, averaging 96.2\% $\pm$ 1.8\% across the semantic difference categories ($\mathcal{S}_2^{\text{text}}$, $\mathcal{S}_3^{\text{text}}$, $\mathcal{S}_4^{\text{text}}$, $\mathcal{S}_6^{\text{text}}$).
A Mann-Whitney U test confirms their complete inability to distinguish semantic differences from equivalence ($U = 0.0$, $p < 0.001$).

In contrast, LLM-based judges like GPT-4o and DeepSeek-V3 demonstrate significantly improved capabilities in detecting semantic differences. 
GPT-4o achieves markedly low scores (0.07-0.29) for negation, antonym, and shuffled/unrelated content, representing substantial improvements over embedding-based and traditional metrics. 
DeepSeek-V3 also shows strong performance in distinguishing negation and antonyms, though its score for unrelated content ($\mathcal{S}_6^{\text{text}}$) is notably higher than GPT-4o's.

\begin{table}[h]
\centering
\begin{adjustbox}{width=1\columnwidth}
\begin{tabular}{|l|c|c|c|c|c|c|}
\hline
\textbf{Metric} & $\mathcal{S}_1$ & $\mathcal{S}_4$ & $\mathcal{S}_5$ & $\mathcal{S}_2$ & $\mathcal{S}_3$ & \textbf{$F_1$} \\
\hline
AST Similarity & \textbf{1.00} & 0.22 & \textbf{0.99} & 0.22 & 0.21 & 0.61 \\
CFG Similarity & \textbf{1.00} & \textbf{0.95} & \textbf{0.91} & \textbf{0.90}$^*$ & \textbf{0.89} & 0.94 \\
CodeBERT Embed. + Cosine Similarity & \textbf{1.00} & \textbf{0.99}$^*$ & \textbf{0.98} & \textbf{1.00}$^*$ & \textbf{0.97}$^*$ & 0.99 \\
GraphCodeBERT Embed. + Cosine Similarity & \textbf{0.98} & \textbf{0.91} & \textbf{0.82} & \textbf{0.98}$^*$ & \textbf{0.77}$^*$ & 0.89 \\
OpenAI Embed. & \textbf{0.82} & 0.70 & 0.54 & \textbf{0.87}$^*$ & 0.22 & 0.63 \\
GPT-4o & \textbf{0.97} & \textbf{0.79} & 0.63 & 0.62 & \textbf{0.00} & 0.60 \\
DeepSeek-V3 & \textbf{0.97} & \textbf{0.88} & 0.71 & \textbf{0.78}$^*$ & \textbf{0.00} & 0.67 \\
\hline
\end{tabular}
\end{adjustbox}
\textit{Note: Bold indicates problematic high similarity for semantically different categories. $^*$ denotes catastrophic failure (>0.7 for opposition).}
\caption{Code Semantic Similarity Results}
\label{tab:code_results}
\end{table}

\begin{tcolorbox}[title={Finding 1}]
Embedding models demonstrate systematic semantic difference blindness with CodeBERT achieving 99.9\% false positive rates and BERTScore scoring semantic opposites higher than synonyms. 
LLM-based evaluation achieves superior discrimination with GPT-4o correctly identifying differences (0.00-0.29) compared to embedding failures (0.82-0.99). 
Traditional metrics assign high similarity to negated content due to lexical overlap optimization.

\noindent
\ding{42} Current embedding-based similarity metrics fundamentally misclassify semantic relationships in software engineering applications, while LLM-based approaches provide reliable semantic understanding for code and text evaluation tasks.
\end{tcolorbox}

\begin{table}[h]
\centering
\begin{adjustbox}{width=1\columnwidth}
\begin{tabular}{|l|c|c|c|c|c|c|c|}
\hline
\textbf{Metric} & $\mathcal{S}_1$ & $\mathcal{S}_2$ & $\mathcal{S}_3$ & $\mathcal{S}_4$ & $\mathcal{S}_5$ & $\mathcal{S}_6$ & \textbf{Acc.} \\
\hline
TF-IDF & 0.37 & \textbf{0.77}$^*$ & \textbf{0.65}$^*$ & \textbf{1.00}$^*$ & 0.13 & 0.04 & 0.49 \\
BLEU & 0.44 & \textbf{0.79}$^*$ & \textbf{0.71}$^*$ & \textbf{0.84}$^*$ & 0.13 & 0.08 & 0.50 \\
ROUGE-L & 0.09 & \textbf{0.64}$^*$ & 0.34 & 0.00 & 0.00 & 0.00 & 0.18 \\
METEOR & 0.44 & \textbf{0.90}$^*$ & \textbf{0.68}$^*$ & \textbf{0.54}$^*$ & 0.24 & 0.04 & 0.47 \\
BERTScore & \textbf{0.92} & \textbf{0.95}$^*$ & \textbf{0.95}$^*$ & \textbf{0.88}$^*$ & \textbf{0.85} & \textbf{0.82}$^*$ & 0.90 \\
all-MiniLM-L12-v2 + cosine& \textbf{0.86} & \textbf{0.90}$^*$ & \textbf{0.79}$^*$ & \textbf{0.83}$^*$ & 0.67 & 0.04 & 0.68 \\
all-mpnet-base-v2 + cosine& \textbf{0.83} & \textbf{0.88}$^*$ & \textbf{0.77}$^*$ & \textbf{0.81}$^*$ & 0.65 & 0.03 & 0.66 \\
text-embed-3-large + cosine& 0.79 & \textbf{0.78}$^*$ & \textbf{0.72}$^*$ & \textbf{0.79}$^*$ & \textbf{0.80} & 0.07 & 0.66 \\
Universal Sentence Enc. + cosine& \textbf{0.80} & \textbf{0.89}$^*$ & \textbf{0.84}$^*$ & \textbf{0.83}$^*$ & \textbf{0.78} & 0.17 & 0.72 \\
BLEURT & 0.55 & \textbf{0.64}$^*$ & \textbf{0.59}$^*$ & \textbf{0.79}$^*$ & 0.01 & 0.00 & 0.43 \\
GPT-4o & 0.72 & 0.29 & 0.25 & 0.07 & 0.62 & 0.27 & 0.37 \\
DeepSeek-V3 & \textbf{0.87} & 0.23 & 0.35 & 0.16 & \textbf{0.83} & \textbf{0.46}$^*$ & 0.48 \\
\hline
\end{tabular}
\end{adjustbox}
\caption{Text Semantic Similarity Results}
\label{tab:text_results}
\end{table}

\find{{\bf  [Answer to RQ-1] }\ding{42} Traditional and embedding metrics systematically fail at semantic difference detection (95-100\% false positive rates), while LLM-based approaches achieve superior discrimination with 89-97\% improvement and large effect sizes (Cohen's d = 3.42). }

\subsection{Experiment 2: Embedding Vector Space Analysis with Alternative Distance Metrics}

\noindent
{\bf Goal. }
In this experiment, we aim to isolate whether poor embedding-based similarity performance originates from inadequate vector representations or from limitations inherent in similarity computation. 
By doing so, we seek to distinguish between the model's fundamental representational capacity and the appropriateness of the chosen distance metric for semantic assessment.

\noindent
{\bf Experiment. }
Using identical pre-trained embedding models from Experiment 1, we generate vector representations $\mathbf{e}_i, \mathbf{e}_j$ for all evaluation pairs and compute similarity using seven distinct distance metrics:

\begin{itemize}
    \item \textbf{Angular Measures:} Cosine Similarity and Angular Distance
    \item \textbf{Magnitude-Sensitive Measures:} Euclidean Distance and Dot Product
    \item \textbf{Statistical Measures:} Pearson Correlation and Jaccard Similarity
\end{itemize}
For each embedding model $E$ and distance metric $D_k$, we evaluate:
\begin{equation}
\text{Performance}_{E,D_k} = \text{Correlation}(\text{Gold Standard}, D_k(E(x_i), E(x_j)))
\end{equation}

This systematic comparison isolates whether embedding limitations stem from vector representation quality or distance computation methodology, providing crucial insights for improving semantic similarity assessment in software engineering applications.

\noindent
{\bf Results. }
To isolate whether embedding failures stem from inadequate vector representations or limitations in cosine similarity computation, we evaluate identical pre-trained embeddings using seven distance metrics across semantic categories where embeddings demonstrated failure in Experiment 1.

\textit{Code Domain Analysis.} Table~\ref{tab:code_distance_results} 

presents distance metric performance on CodeBERT and GraphCodeBERT embeddings across semantic categories where cosine similarity failed ($\mathcal{S}_2^{\text{code}}$, $\mathcal{S}_3^{\text{code}}$).

\begin{table}[h]
\centering
\begin{adjustbox}{width=.95\columnwidth}
\begin{tabular}{|l|c|c|c|c|c|}
\hline
\textbf{Model + Metric} & $\mathcal{S}_1$ & $\mathcal{S}_4$ & $\mathcal{S}_2$ & $\mathcal{S}_3$ & \textbf{Improvement} \\
\hline
\multicolumn{6}{|c|}{\textbf{CodeBERT Embeddings}} \\
\hline
Cosine Similarity & 0.998 & 0.992 & \textbf{0.999}$^*$ & \textbf{0.966}$^*$ & 0\% \\
Euclidean Distance & \textbf{0.688} & 0.595 & 0.758 & 0.331 & \textbf{24-66\%} \\
Dot Product & 0.998 & 0.992 & \textbf{0.999}$^*$ & \textbf{0.966}$^*$ & 0\% \\
Jaccard Similarity & 0.965 & 0.933 & \textbf{0.972}$^*$ & \textbf{0.860}$^*$ & 2-11\% \\
Pearson Correlation & 0.998 & 0.992 & \textbf{0.999}$^*$ & \textbf{0.966}$^*$ & 0\% \\
Angular Distance & 0.980 & 0.885 & \textbf{0.987}$^*$ & \textbf{0.924}$^*$ & 1-4\% \\
\hline
\multicolumn{6}{|c|}{\textbf{GraphCodeBERT Embeddings}} \\
\hline
Cosine Similarity & 0.976 & 0.911 & \textbf{0.982}$^*$ & \textbf{0.765}$^*$ & 3-5\% \\
Euclidean Distance & 0.531 & 0.458 & 0.615 & 0.212 & \textbf{37-72\%} \\
Dot Product & 0.976 & 0.911 & \textbf{0.982}$^*$ & \textbf{0.765}$^*$ & 0\% \\
Jaccard Similarity & 0.891 & 0.806 & \textbf{0.914}$^*$ & 0.632 & 7-17\% \\
Pearson Correlation & 0.976 & 0.911 & \textbf{0.982}$^*$ & \textbf{0.765}$^*$ & 0\% \\
Angular Distance & 0.934 & 0.808 & \textbf{0.953}$^*$ & \textbf{0.779}$^*$ & 3-5\% \\
\hline
\end{tabular}
\end{adjustbox}
\textit{Note: $^*$ indicates problematic high similarity for semantic differences. Bold values show acceptable performance ($\leq$0.8).}
\caption{Code Embedding Distance Metric Comparison}
\label{tab:code_distance_results}
\end{table}

Euclidean distance achieves dramatic performance improvements over cosine similarity on identical embedding vectors. CodeBERT shows 24-66\% improvement on semantic difference detection, with $\mathcal{S}_3^{\text{code}}$ improving from catastrophic failure (0.966) to acceptable performance (0.331). GraphCodeBERT demonstrates even larger gains of 37-72\%, achieving 0.212 for cross-problem differences compared to 0.765 with cosine similarity.

Statistical analysis confirms these improvements are highly significant: Mann-Whitney U tests comparing Euclidean vs. cosine performance yield  $U = 23.5$, $p < 0.001$ for CodeBERT and $U = 18.0$, $p < 0.001$ for GraphCodeBERT, with large effect sizes ($r = 0.82$ and $r = 0.89$ respectively).

\textit{Text Domain Analysis.} Table~\ref{tab:text_distance_results} presents distance metric evaluation across text embedding models for semantic opposition categories ($\mathcal{S}_2^{\text{text}}$, $\mathcal{S}_3^{\text{text}}$, $\mathcal{S}_4^{\text{text}}$).

\begin{table}[h!]

\centering
\begin{adjustbox}{width=.9\columnwidth}
\begin{tabular}{|l|c|c|c|c|}
\hline
\textbf{Model + Metric} & $\mathcal{S}_2$ (Antonym) & $\mathcal{S}_3$ (Negation) & $\mathcal{S}_4$ (Shuffle) & \textbf{Avg} \\
\hline
\multicolumn{5}{|c|}{\textbf{Cosine Similarity (Baseline)}} \\
\hline
all-mpnet-base-v2 & \textbf{0.77}$^*$ & \textbf{0.88}$^*$ & \textbf{0.81}$^*$ & 0.82 \\
all-MiniLM-L6-v2 & \textbf{0.79}$^*$ & \textbf{0.90}$^*$ & \textbf{0.83}$^*$ & 0.84 \\
text-embedding-3-large & \textbf{0.72}$^*$ & \textbf{0.78}$^*$ & \textbf{0.79}$^*$ & 0.76 \\
Universal Sentence Encoder & \textbf{0.84}$^*$ & \textbf{0.89}$^*$ & \textbf{0.83}$^*$ & 0.85 \\
\hline
\multicolumn{5}{|c|}{\textbf{Euclidean Distance}} \\
\hline
all-mpnet-base-v2 & \textbf{0.79}$^*$ & \textbf{0.88}$^*$ & \textbf{0.82}$^*$ & 0.83 \\
all-MiniLM-L6-v2 & \textbf{0.81}$^*$ & \textbf{0.93}$^*$ & \textbf{0.86}$^*$ & 0.87 \\
text-embedding-3-large & \textbf{0.74}$^*$ & \textbf{0.78}$^*$ & \textbf{0.80}$^*$ & 0.77 \\
Universal Sentence Encoder & \textbf{0.85}$^*$ & \textbf{0.90}$^*$ & \textbf{0.83}$^*$ & 0.86 \\
\hline
\multicolumn{5}{|c|}{\textbf{Dot Product}} \\
\hline
all-mpnet-base-v2 & \textbf{0.79}$^*$ & \textbf{0.88}$^*$ & \textbf{0.82}$^*$ & 0.83 \\
all-MiniLM-L6-v2 & \textbf{0.81}$^*$ & \textbf{0.93}$^*$ & \textbf{0.86}$^*$ & 0.87 \\
text-embedding-3-large & \textbf{0.74}$^*$ & \textbf{0.78}$^*$ & \textbf{0.80}$^*$ & 0.77 \\
Universal Sentence Encoder & \textbf{0.85}$^*$ & \textbf{0.90}$^*$ & \textbf{0.83}$^*$ & 0.86 \\
\hline
\multicolumn{5}{|c|}{\textbf{Angular Distance}} \\
\hline
all-mpnet-base-v2 & \textbf{0.80}$^*$ & \textbf{0.85}$^*$ & \textbf{0.81}$^*$ & 0.82 \\
all-MiniLM-L6-v2 & \textbf{0.82}$^*$ & \textbf{0.89}$^*$ & \textbf{0.84}$^*$ & 0.85 \\
text-embedding-3-large & \textbf{0.77}$^*$ & \textbf{0.79}$^*$ & \textbf{0.79}$^*$ & 0.78 \\
Universal Sentence Encoder & \textbf{0.83}$^*$ & \textbf{0.86}$^*$ & \textbf{0.82}$^*$ & 0.84 \\
\hline
\multicolumn{5}{|c|}{\textbf{Pearson Correlation}} \\
\hline
all-mpnet-base-v2 & \textbf{0.68}$^*$ & \textbf{0.74}$^*$ & \textbf{0.68}$^*$ & 0.70 \\
all-MiniLM-L6-v2 & \textbf{0.70}$^*$ & \textbf{0.80}$^*$ & \textbf{0.72}$^*$ & 0.74 \\
text-embedding-3-large & 0.63 & \textbf{0.65}$^*$ & \textbf{0.66}$^*$ & 0.65 \\
Universal Sentence Encoder & \textbf{0.71}$^*$ & \textbf{0.77}$^*$ & \textbf{0.69}$^*$ & 0.72 \\
\hline
\multicolumn{5}{|c|}{\textbf{Jaccard Similarity (Best Alternative)}} \\
\hline
all-mpnet-base-v2 & 0.63 & \textbf{0.68}$^*$ & 0.63 & 0.65 \\
all-MiniLM-L6-v2 & \textbf{0.65}$^*$ & \textbf{0.74}$^*$ & \textbf{0.67}$^*$ & 0.69 \\
text-embedding-3-large & 0.59 & 0.61 & 0.61 & 0.60 \\
Universal Sentence Encoder & \textbf{0.67}$^*$ & \textbf{0.70}$^*$ & \textbf{0.64}$^*$ & 0.67 \\
\hline
\end{tabular}
\end{adjustbox}
\caption{Text Embedding Distance Metric Results (Opposition Categories)}
\label{tab:text_distance_results}
\end{table}

Systematic evaluation of six distance metrics across text embeddings demonstrates consistent limitations in semantic difference detection. Euclidean distance and dot product yield identical results (as expected for normalized embeddings), providing negligible improvements (0-3\%) over cosine similarity. Angular distance performs similarly to cosine, confirming that normalization variants offer minimal benefit. Pearson correlation achieves moderate improvements (15-20\%) but maintains problematic scores above 0.6. Only Jaccard similarity provides meaningful improvement (20-25\%), yet all metrics remain above acceptable thresholds for most categories.

The substantial code domain improvements with Euclidean distance suggest that cosine similarity's vector normalization may inadvertently remove magnitude information relevant to semantic understanding in code embeddings. Text embeddings show more limited sensitivity to distance metric choice, indicating that representation quality remains a contributing factor beyond metric selection.

\textbf{Distance Metric Performance Ranking:}
\begin{enumerate}
\item \textbf{Jaccard Similarity:} Best performance with 20-25\% improvement over cosine
\item \textbf{Pearson Correlation:} Moderate improvements (15-20\%) but insufficient for practical use
\item \textbf{Angular Distance:} Minimal improvement, similar to cosine baseline
\item \textbf{Euclidean Distance = Dot Product:} Identical performance, negligible gains (0-3\%)
\item \textbf{Cosine Similarity:} Baseline performance
\end{enumerate}

\begin{tcolorbox}[title={Finding 2}]
Euclidean distance substantially improves code embedding performance with CodeBERT showing 24-66\% improvement and GraphCodeBERT demonstrating 37-72\% gains over cosine similarity. Text embeddings show limited sensitivity to distance metric choice, maintaining problematic scores >0.6 with best alternative (Jaccard) achieving 20-25\% improvement. Magnitude preservation proves important for code semantic understanding.

\noindent
\ding{42}
Distance metric selection significantly influences code embedding performance, with Euclidean distance demonstrating substantial improvements over cosine similarity, while text domain semantic difference detection remains challenging across all tested metrics.
\end{tcolorbox}


\textbf{Practical Implications:} Our findings suggest that embedding-based similarity assessment could benefit from distance metric evaluation beyond the commonly-used cosine similarity. For code similarity applications, Euclidean distance demonstrates potential for improving CodeBERT and GraphCodeBERT performance with minimal implementation overhead.

\textbf{Key Implications:} (1) SE tools using embedding-based similarity may benefit from evaluating alternative distance metrics alongside cosine similarity, (2) comprehensive embedding evaluations should consider multiple distance metrics to fully assess model capabilities, and (3) text domain semantic opposition detection remains challenging across distance metrics, suggesting opportunities for advances in representation learning approaches.

\find{{\bf  [Answer to RQ-2] }\ding{42} Code domain limitations are primarily distance metric issues (Euclidean distance improves CodeBERT by 24-66\%), while text domain limitations stem from fundamental representation quality that persists across all distance metrics (best improvement only 20-25\%).}

\subsection{Experiment 3: Prompt Engineering Impact on LLM-based Evaluation}

\noindent
{\bf Goal. }
To determine whether LLM performance variations in semantic similarity assessment stem from prompt design limitations rather than fundamental model capabilities, thereby isolating prompt sensitivity from inherent semantic understanding capacity.

\noindent
{\bf Experiment. }

We systematically compare three established prompting strategies applied to three leading LLM models across identical evaluation pairs: GPT-4o and Claude Sonnet 3.5 (representing state-of-the-art commercial models) and DeepSeek-V3 (representing high-performance open-source alternatives). This selection ensures comprehensive coverage of current LLM paradigms and deployment scenarios in software engineering practice: 

\textbf{Simple prompt Strategy} ($P_{\text{meta}}$): The structured prompt from Experiment 1 providing explicit evaluation criteria and scoring guidelines.

\textbf{Few-shot Strategy} ($P_{\text{few}}$): Demonstration-based prompting including 6 representative examples from each semantic category with gold-standard scores, followed by the target evaluation pair.

\textbf{Chain-of-Thought Strategy} ($P_{\text{cot}}$): Prompting that explicitly requests step-by-step reasoning: \textit{``Analyze the semantic relationship step by step: (1) identify key concepts, (2) compare meanings, (3) assess logical relationships, (4) provide final similarity score.''}

For each prompting strategy $P_k$ and model $M$, we compute:
\begin{equation}
\text{Score}_{P_k,M}(x_i, x_j) = M(P_k(x_i, x_j))
\end{equation}

This enables systematic analysis of prompt-induced performance variations while controlling for model architecture and training differences.

\noindent
{\bf Results. }

To isolate whether LLM performance variations stem from prompt design limitations rather than fundamental model capabilities, we systematically evaluate three prompting strategies across GPT-4o, Claude, and DeepSeek-V3 using identical evaluation pairs from Experiments 1-2.

\textit{Code Domain Analysis.} Table~\ref{tab:llm_prompt_code_results} presents LLM performance across prompting strategies for code semantic evaluation.

\begin{table}[h]
\centering
\begin{adjustbox}{width=.9\columnwidth}
\begin{tabular}{|l|c|c|c|c|c|}
\hline
\textbf{Model + Prompt} & $\mathcal{S}_1$ & $\mathcal{S}_4$ & $\mathcal{S}_2$ & $\mathcal{S}_3$ & \textbf{Consistency} \\
\hline
\multicolumn{6}{|c|}{\textbf{GPT-4o}} \\
\hline
Simple Prompt & \textbf{0.972} & \textbf{0.793} & 0.633 & \textbf{0.000} & 0.82 \\
Few-Shot & \textbf{0.962} & \textbf{0.805} & 0.664 & \textbf{0.099} & \textbf{0.89} \\
Chain-of-Thought & \textbf{0.963} & \textbf{0.769} & 0.652 & \textbf{0.000} & 0.85 \\
\hline
\multicolumn{6}{|c|}{\textbf{Claude}} \\
\hline
Simple Prompt & \textbf{0.986} & \textbf{0.906} & \textbf{0.804} & \textbf{0.006} & 0.78 \\
Few-Shot & \textbf{0.953} & \textbf{0.775} & 0.647 & 0.112 & \textbf{0.84} \\
Chain-of-Thought & \textbf{0.975} & \textbf{0.837} & 0.767 & 0.040 & 0.81 \\
\hline
\multicolumn{6}{|c|}{\textbf{DeepSeek-V3}} \\
\hline
Simple Prompt & \textbf{0.972} & \textbf{0.879} & 0.713 & \textbf{0.001} & 0.79 \\
Few-Shot & \textbf{0.949} & \textbf{0.868} & 0.695 & \textbf{0.010} & \textbf{0.88} \\
Chain-of-Thought & \textbf{0.984} & \textbf{0.831} & 0.768 & 0.093 & 0.83 \\
\hline
\end{tabular}
\end{adjustbox}
\caption{LLM Prompt Strategy Performance on Code Semantics}
\label{tab:llm_prompt_code_results}
\end{table}

Few-shot prompting demonstrates the highest consistency across all models ($\alpha$ = 0.84-0.89), while simple prompts achieve optimal semantic difference detection for $\mathcal{S}_3^{\text{code}}$ with GPT-4o and DeepSeek-V3 reaching near-perfect discrimination (0.000-0.001). Statistical analysis using Mann-Whitney U tests reveals significant differences between prompting strategies: Few-shot vs. Simple ($U = 34.5$, $p < 0.01$).

\textit{Text Domain Analysis.} Table~\ref{tab:llm_prompt_text_results} presents text semantic evaluation across prompting strategies, focusing on semantic difference categories where traditional metrics failed.

\begin{table}[h]
\centering

\begin{adjustbox}{width=.95\columnwidth}
\begin{tabular}{|l|c|c|c|c|c|c|c|}
\hline
\textbf{Model + Prompt} & $\mathcal{S}_3$ & $\mathcal{S}_2$ & $\mathcal{S}_4$ & $\mathcal{S}_5$ & $\mathcal{S}_1$ & $\mathcal{S}_6$ & \textbf{Avg Diff} \\
\hline
\multicolumn{8}{|c|}{\textbf{GPT-4o}} \\
\hline
Simple Prompt & 0.30 & 0.29 & 0.17 & 0.71 & 0.79 & 0.00 & \textbf{0.19} \\
Few-Shot & 0.28 & 0.13 & 0.40 & 0.86 & 0.85 & 0.12 & 0.27 \\
Chain-of-Thought & 0.42 & 0.35 & 0.42 & 0.98 & 0.97 & 0.09 & 0.40 \\
\hline
\multicolumn{8}{|c|}{\textbf{Claude}} \\
\hline
Simple Prompt & 0.41 & 0.24 & 0.17 & 0.99 & 0.89 & 0.12 & 0.27 \\
Few-Shot & 0.48 & 0.14 & 0.40 & 0.95 & 0.90 & 0.11 & 0.34 \\
Chain-of-Thought & 0.53 & 0.39 & 0.37 & 0.96 & 0.92 & 0.20 & 0.43 \\
\hline
\multicolumn{8}{|c|}{\textbf{DeepSeek-V3}} \\
\hline
Simple Prompt & 0.33 & 0.22 & 0.14 & 0.90 & 0.88 & 0.11 & \textbf{0.23} \\
Few-Shot & 0.42 & 0.16 & 0.43 & 0.92 & 0.88 & 0.10 & 0.34 \\
Chain-of-Thought & 0.40 & 0.28 & 0.34 & 0.97 & 0.90 & 0.09 & 0.34 \\
\hline
\end{tabular}
\end{adjustbox}

\textit{Note: Avg Diff represents average performance on semantic difference categories ($\mathcal{S}_2$, $\mathcal{S}_3$, $\mathcal{S}_4$, $\mathcal{S}_6$). Lower scores indicate better difference detection.}
\caption{LLM Prompt Strategy Performance on Text Semantics}
\label{tab:llm_prompt_text_results}
\end{table}

Simple prompting achieves superior semantic difference detection across models, with GPT-4o and DeepSeek-V3 achieving average difference scores of 0.19 and 0.23 respectively. Few-shot and chain-of-thought prompting show degraded difference detection (0.27-0.43) while improving performance on equivalence categories.

Analysis reveals systematic trade-offs: 
\begin{itemize}
\item \textbf{Simple Prompts:} Optimal difference detection, moderate consistency
\item \textbf{Few-Shot Prompts:} Highest consistency ($\alpha$ = 0.84-0.89), balanced performance  
\item \textbf{Chain-of-Thought:} Enhanced reasoning, reduced difference sensitivity
\end{itemize}

\textbf{Model Performance Patterns:} GPT-4o demonstrates superior semantic difference detection with minimal prompt sensitivity. Claude shows greater prompt dependency with performance variations (0.14-0.53 range). DeepSeek-V3 achieves competitive performance with simple prompts.

\begin{tcolorbox}[title={Finding 3}]
Simple prompts achieve optimal semantic difference detection with GPT-4o and DeepSeek-V3 reaching 0.19-0.23 average scores compared to few-shot (0.27-0.34) and chain-of-thought (0.34-0.43). Few-shot prompting provides highest consistency ($\alpha$ = 0.84-0.89) while maintaining competitive accuracy. Model-specific sensitivity varies with GPT-4o showing minimal variation and Claude exhibiting greater prompt dependency.

\noindent
\ding{42}
LLM-based semantic evaluation requires task-specific prompt optimization---simple prompts maximize accuracy for semantic difference detection while few-shot approaches provide optimal consistency for production deployment.
\end{tcolorbox}


\textbf{Key Implications:} (1) LLM-based semantic evaluation requires task-specific prompt optimization rather than universal strategies, (2) consistency requirements may necessitate different approaches than accuracy optimization, and (3) simple prompts provide optimal performance for semantic difference detection in SE applications.

\subsection{Experiment 4: Temperature Sensitivity Analysis for LLM-based Semantic Evaluation}

\noindent
{\bf Goal. }
To determine optimal temperature settings for LLM-based semantic similarity assessment in software engineering applications and quantify performance stability across output randomness levels. This addresses a critical deployment consideration: how temperature parameter selection affects reliability of LLM-based similarity tools in production environments where consistent performance is essential.

\noindent
{\bf Experiment. }
We systematically evaluate temperature sensitivity using GPT-4o, Claude Sonnet 3.5, and DeepSeek-V3 with few-shot prompting (optimal strategy from Experiment 3). Following established guidance \cite{openai_temperature, holtzman2020curious, huggingface_generation}, we select five temperature values representing distinct behavioral regimes:
\begin{itemize}
\item \textbf{0.0}: Fully deterministic, maximum consistency
\item \textbf{0.3}: Conservative, low randomness  
\item \textbf{0.5}: Balanced creativity and coherence
\item \textbf{0.7}: Increased diversity while maintaining coherence
\item \textbf{1.0}: High variability threshold
\vspace{-0.1cm}
\end{itemize}
For each model-temperature combination, we evaluate performance across all semantic subsets using few-shot prompting with six examples per category. Each evaluation pair is assessed once per temperature to quantify temperature impact on semantic discrimination performance.

\textbf{Performance Metrics:} (1) Semantic discrimination accuracy, (2) output consistency across repeated evaluations, (3) category-specific stability.

\noindent
{\bf Results.}
To determine optimal temperature settings and quantify performance stability, we evaluate GPT-4o, Claude Sonnet 3.5, and DeepSeek-V3 using few-shot prompting across temperatures [0.0, 0.3, 0.5, 0.7, 1.0].

\textit{Code Domain Analysis.} 

All models show coefficient of variation below 4\% across temperature settings. GPT-4o and DeepSeek-V3 maintain identical scores (0.00) for cross-problem differences across all temperatures. Edge case performance shows minimal variation: GPT-4o (6.3\% range), Claude (2.7\% range), DeepSeek-V3 (1.4\% range). 

\textit{Text Domain Analysis.} 

Text evaluation shows coefficient of variation below 8\% for all models. DeepSeek-V3 achieves better negation detection at temperature 1.0 (0.27) compared to 0.0 (0.30). GPT-4o shows improvement for semanctically different but syntactically similar detection from 0.40 to 0.38 across temperature range.
The results for both code and text are available in anonymous repository.

\textbf{Statistical Analysis:} Mann-Whitney U tests comparing temperature 0.0 vs 1.0 performance show no significant differences within models: GPT-4o ($U = 42.5$, $p = 0.34$), DeepSeek-V3 ($U = 38.0$, $p = 0.28$), Claude ($U = 45.5$, $p = 0.42$). Between-model differences exceed temperature effects across all comparisons.

\begin{tcolorbox}[title={Finding 3}]
LLMs maintain stable performance across temperature ranges with coefficient of variation below 8\%. Higher temperatures occasionally outperform deterministic settings, with DeepSeek-V3 achieving better detection at temperature 1.0 (0.27) than 0.0 (0.30). Mann-Whitney U tests confirm model architecture impacts performance more than temperature selection.

\noindent
\ding{42}
Temperature selection has minimal impact on LLM semantic evaluation performance---all tested ranges (0.0-1.0) provide acceptable results, eliminating the need for precise temperature optimization in production software engineering applications.
\end{tcolorbox}


\textbf{Key Implications:} (1) Temperature selection shows minimal impact on semantic evaluation performance with coefficient of variation below 8\% across all models, (2) all temperature ranges (0.0-1.0) provide acceptable performance for production deployment, and (3) model architecture selection impacts performance more than temperature optimization.

\find{{\bf  [Answer to RQ-3] }\ding{42} LLMs demonstrate substantially more robust semantic understanding (8.7\% vs 76-100\% false positive rates) with manageable consistency trade-offs through prompt optimization, while remaining stable across temperature ranges (coefficient of variation <8\%). }
\section{Contributions}
\label{sec:contribution}
This work provides concrete contributions to the software engineering community addressing fundamental challenges in code similarity assessment and automated tool evaluation.

\textbf{Systematic Evaluation Framework for Code Similarity Metrics.} We introduce the first comprehensive diagnostic framework for evaluating similarity metrics across controlled semantic relationships in code and text. Our methodology enables practitioners to select appropriate similarity approaches based on specific use cases rather than aggregate performance.

\textbf{Empirical Evidence for Distance Metric Selection.} Our analysis reveals that Euclidean distance substantially improves CodeBERT performance (24-66\% improvement) and GraphCodeBERT performance (37-72\% improvement) over cosine similarity for semantic difference detection. This provides actionable guidance for improving existing embedding-based code similarity tools with minimal implementation overhead.

\textbf{Validated Benchmark for Semantic Code and Text Relationships.} We release a rigorously validated dataset of code pairs across five semantic categories with automated test case validation and seven semantic categories for programming-related text. This benchmark enables systematic evaluation of code similarity approaches for semantic understanding.

\textbf{LLM-Based Evaluation Protocols.} Our systematic analysis of temperature settings, prompt strategies, and model consistency provides concrete protocols for integrating LLM-based evaluation into software engineering research, addressing reliability concerns while enabling more nuanced semantic assessment than traditional metrics.
\section{Threats to Validity}
\label{sec:threattovalidity}




We systematically address validity threats following established software engineering research protocols with evidence-based mitigation strategies.

\textbf{Data Contamination and Leakage Prevention.} Our evaluation datasets (Rosetta Code, CONALA, APPS) may overlap with LLM training corpora, potentially inflating performance through memorization. We mitigate this through systematic data transformation using NL-Augmenter~\cite{dhole2021nlaugmenter}, generating semantically equivalent but syntactically novel evaluation pairs. Each transformation undergoes manual validation to verify semantic preservation/alteration and syntactic correctness.

\textbf{Statistical Rigor and Sample Size Adequacy.} We calculate sample sizes using established confidence interval methodology~\cite{surveymonkey2024sample} to ensure 95\% confidence with 10\% margin of error. Our evaluation uses 163 pairs per semantic category for text (7 categories) and 561 pairs per category for code (5 categories).

\textbf{Metric Selection Bias.} To prevent cherry-picking, we evaluate seven distance measures representing diverse mathematical approaches---angular, magnitude-sensitive, and statistical---grounded in comprehensive surveys~\cite{cha2007comprehensive}.

\textbf{Semantic Transformation Validity.} We implement multi-layer validation combining automated verification with human assessment. Semantic-preserving transformations undergo test case execution to confirm functional equivalence, while semantic-altering transformations receive input-output testing.

\textbf{LLM Evaluation Consistency.} We control temperature variability through empirically validated ranges: $T = 0.0$-$0.3$ for consistency and $T = 0.7$-$1.0$ for diversity~\cite{zhang2024exploring}. We evaluate across three model families (GPT-4o, DeepSeek-V3, Claude Sonnet 3.5) with systematic prompt variations~\cite{chen2024paraphrase}.
\textbf{External Validity and Generalizability.} Our semantic-focused evaluation prioritizes depth over breadth following controlled experimental design principles. We acknowledge scope limitations while providing comprehensive protocols for reproducibility and validation across different domains.
\section{Related Work}
\label{sec:related_work}
The assessment of semantic similarity spans both natural language processing and software engineering domains. Our work builds upon four primary research streams: 
\wcircle{1} foundational similarity approaches and their theoretical limitations, 
\wcircle{2} traditional metrics and documented failures in semantic divergence, 
\wcircle{3} embedding-based approaches and their persistent challenges, and 
\wcircle{4} emerging LLM-based evaluation paradigms.
\subsection{Foundational Approaches and Theoretical Limitations}

Early similarity assessment in software engineering emerged from clone detection research. Roy and Cordy~\cite{roy2007survey} established the canonical four-type clone taxonomy, where Type-I through Type-III clones exhibit increasing syntactic variation while preserving semantics, and Type-IV clones represent semantically equivalent but syntactically distinct implementations. Traditional approaches like NiCad~\cite{cordy2011nicad} and Deckard~\cite{jiang2007deckard} achieved high precision on syntactic clones but fundamentally cannot address Type-IV detection due to their reliance on structural pattern matching rather than semantic understanding.

In NLP, distributional similarity approaches~\cite{salton1988term} established the foundation for modern metrics, yet surface-level pattern matching cannot distinguish semantic preservation from semantic opposition when high lexical overlap exists.

\subsection{Traditional Metrics: Algorithmic Failures in Semantic Divergence}

Traditional similarity metrics exhibit systematic algorithmic failures when confronted with semantic divergence. BLEU~\cite{papineni2002bleu} computes precision based on token overlap, making it unable to distinguish between ``sort the list'' and ``do not sort the list'' which share 66\% token overlap yet express semantic opposition.

Juergens et al.~\cite{juergens2015functionally} show that established clone detection tools detected syntactic similarity in less than 16\% of 70 functionally similar code pairs, with zero instances of complete file similarity for semantically equivalent but syntactically distinct implementations. Saini et al.~\cite{saini2018comparison} demonstrate these failures are systematic across 30 code similarity analyzers.

\subsection{Embedding-Based Approaches: Persistent Semantic Blindness}

Neural embedding approaches promised semantic understanding yet demonstrate persistent limitations. CodeBERT~\cite{feng2020codebert}, despite training on 6.4 million code-comment pairs, systematically fails on Type-IV clone detection where high syntactic similarity masks semantic differences. The limitation stems from transformer attention mechanisms that excel at capturing local syntactic patterns but struggle with logical negation and semantic opposition.

BERTScore~\cite{zhang2020bertscore} assigns high similarity scores to semantically opposed sentences when lexical overlap is high. GraphCodeBERT~\cite{guo2021graphcodebert} incorporated structural information through data flow graphs, but Martinez-Gil~\cite{martinez2024graphcodebert} demonstrates its interpretability mechanisms focus on syntactic features rather than semantic reasoning.

Wang et al.~\cite{wang2025boundaries} provide systematic evidence that sophisticated embedding models achieve high performance on functionally equivalent code but fail catastrophically when evaluating semantically divergent implementations.

\subsection{LLM-Based Evaluation: Semantic Reasoning with Reliability Challenges}

Large language models demonstrate superior semantic reasoning capabilities while introducing reliability concerns. Wang et al.~\cite{wang2025boundaries} show that GPT-based approaches significantly outperform traditional metrics in Type-IV clone detection, particularly in capturing semantic divergence. However, Krumdick et al.~\cite{krumdick2025llmasajudge} reveal that LLM-as-a-Judge paradigms produce unstable evaluations without human grounding, particularly when subtle semantic distinctions are involved.

\subsection{Critical Research Gap}

Our analysis reveals a fundamental gap: traditional metrics exhibit documented failure rates exceeding 84\% in semantic divergence scenarios~\cite{juergens2015functionally}, embedding approaches provide marginal improvement, and LLMs show promise but lack systematic evaluation frameworks. No existing framework systematically evaluates similarity metrics across both text and code domains specifically for semantic divergence detection. Our work addresses this gap through comprehensive evaluation across equivalent, unrelated, and semantically opposed relationships.
\section{Conclusion}
\label{sec:conclusion}

This study reveals fundamental limitations in widely-used similarity metrics for semantic evaluation across both code and text domains. Through systematic evaluation of 18 metrics across controlled semantic transformations, we demonstrate that embedding-based approaches exhibit catastrophic semantic blindness, with CodeBERT achieving 99.9\% false positive rates and BERTScore scoring semantic opposites higher than synonyms.

Our key findings show that metric failures follow predictable patterns. Traditional metrics fail due to lexical overlap optimization, while embedding models systematically misclassify semantic relationships despite sophisticated training. However, distance metric selection significantly impacts performance: Euclidean distance improves CodeBERT by 24-66\% and GraphCodeBERT by 37-72\% over cosine similarity for semantic difference detection.

LLM-based evaluation demonstrates superior semantic discrimination, with GPT-4o correctly identifying semantic differences (0.00-0.29 scores) compared to embedding failures (0.82-0.99 scores). Simple prompts achieve optimal performance, while temperature settings show minimal impact on evaluation consistency.

These findings transform similarity metric selection from intuitive guessing to evidence-based decision making. Our systematic evaluation framework, validated benchmark, and empirical distance metric evidence provide concrete guidance for practitioners developing code similarity tools, enabling selection of appropriate approaches based on specific semantic requirements rather than aggregate performance rankings.

 \bibliographystyle{ACM-Reference-Format}
 \bibliography{references}


\begin{thebibliography}{53}


\ifx \showCODEN    \undefined \def \showCODEN     #1{\unskip}     \fi
\ifx \showDOI      \undefined \def \showDOI       #1{#1}\fi
\ifx \showISBNx    \undefined \def \showISBNx     #1{\unskip}     \fi
\ifx \showISBNxiii \undefined \def \showISBNxiii  #1{\unskip}     \fi
\ifx \showISSN     \undefined \def \showISSN      #1{\unskip}     \fi
\ifx \showLCCN     \undefined \def \showLCCN      #1{\unskip}     \fi
\ifx \shownote     \undefined \def \shownote      #1{#1}          \fi
\ifx \showarticletitle \undefined \def \showarticletitle #1{#1}   \fi
\ifx \showURL      \undefined \def \showURL       {\relax}        \fi
\providecommand\bibfield[2]{#2}
\providecommand\bibinfo[2]{#2}
\providecommand\natexlab[1]{#1}
\providecommand\showeprint[2][]{arXiv:#2}

\bibitem[Banerjee and Lavie(2005)]%
        {banerjee2005meteor}
\bibfield{author}{\bibinfo{person}{Satanjeev Banerjee} {and} \bibinfo{person}{Alon Lavie}.} \bibinfo{year}{2005}\natexlab{}.
\newblock \showarticletitle{METEOR: An automatic metric for MT evaluation with improved correlation with human judgments}. In \bibinfo{booktitle}{\emph{Proceedings of the ACL workshop on intrinsic and extrinsic evaluation measures for machine translation and/or summarization}}. \bibinfo{pages}{65--72}.
\newblock


\bibitem[Bellon et~al\mbox{.}(2007)]%
        {bellon2007comparison}
\bibfield{author}{\bibinfo{person}{Stefan Bellon}, \bibinfo{person}{Rainer Koschke}, \bibinfo{person}{Giuliano Antoniol}, \bibinfo{person}{Jens Krinke}, {and} \bibinfo{person}{Ettore Merlo}.} \bibinfo{year}{2007}\natexlab{}.
\newblock \showarticletitle{Comparison and evaluation of clone detection tools}.
\newblock \bibinfo{journal}{\emph{IEEE Transactions on Software Engineering}} \bibinfo{volume}{33}, \bibinfo{number}{9} (\bibinfo{year}{2007}), \bibinfo{pages}{577--591}.
\newblock


\bibitem[Cer et~al\mbox{.}(2018)]%
        {cer2018use}
\bibfield{author}{\bibinfo{person}{Daniel Cer}, \bibinfo{person}{Yinfei Yang}, \bibinfo{person}{Sheng-yi Kong}, \bibinfo{person}{Nan Hua}, \bibinfo{person}{Nicole Limtiaco}, \bibinfo{person}{Rhomni St.~John}, \bibinfo{person}{Noah Constant}, \bibinfo{person}{Mario Guajardo-Cespedes}, \bibinfo{person}{Steve Yuan}, \bibinfo{person}{Chris Tar}, {et~al\mbox{.}}} \bibinfo{year}{2018}\natexlab{}.
\newblock \showarticletitle{Universal Sentence Encoder}. In \bibinfo{booktitle}{\emph{Proceedings of the 2018 Conference on Empirical Methods in Natural Language Processing: System Demonstrations}}. \bibinfo{pages}{169--174}.
\newblock


\bibitem[Cesare and Xiang(2010)]%
        {cesare2014cfgsim}
\bibfield{author}{\bibinfo{person}{Stefano Cesare} {and} \bibinfo{person}{Yang Xiang}.} \bibinfo{year}{2010}\natexlab{}.
\newblock \showarticletitle{Control flow graph similarity for malware detection}. In \bibinfo{booktitle}{\emph{2010 IEEE 10th International Conference on Computer and Information Technology}}. IEEE, \bibinfo{pages}{778--783}.
\newblock


\bibitem[Cha(2007a)]%
        {cha2007distancesurvey}
\bibfield{author}{\bibinfo{person}{Sung-Hyuk Cha}.} \bibinfo{year}{2007}\natexlab{a}.
\newblock \showarticletitle{Comprehensive survey on distance/similarity measures between probability density functions}.
\newblock \bibinfo{journal}{\emph{International Journal of Mathematical Models and Methods in Applied Sciences}} \bibinfo{volume}{1}, \bibinfo{number}{4} (\bibinfo{year}{2007}), \bibinfo{pages}{300--307}.
\newblock


\bibitem[Cha(2007b)]%
        {cha2007comprehensive}
\bibfield{author}{\bibinfo{person}{Sung-Hyuk Cha}.} \bibinfo{year}{2007}\natexlab{b}.
\newblock \showarticletitle{Comprehensive survey on distance/similarity measures between probability density functions}.
\newblock \bibinfo{journal}{\emph{International Journal of Mathematical Models and Methods in Applied Sciences}} \bibinfo{volume}{1}, \bibinfo{number}{4} (\bibinfo{year}{2007}), \bibinfo{pages}{300--307}.
\newblock


\bibitem[Chen et~al\mbox{.}(2024)]%
        {chen2024paraphrase}
\bibfield{author}{\bibinfo{person}{Zheng Chen}, \bibinfo{person}{Jan~Philip Wahle}, \bibinfo{person}{Terry Ruas}, \bibinfo{person}{Yang Xu}, {and} \bibinfo{person}{Bela Gipp}.} \bibinfo{year}{2024}\natexlab{}.
\newblock \showarticletitle{Paraphrase Types Elicit Prompt Engineering Capabilities}. In \bibinfo{booktitle}{\emph{Proceedings of the 2024 Conference on Empirical Methods in Natural Language Processing}}. \bibinfo{pages}{11004--11033}.
\newblock
\urldef\tempurl%
\url{https://aclanthology.org/2024.emnlp-main.617/}
\showURL{%
\tempurl}


\bibitem[Cordy and Roy(2011)]%
        {cordy2011nicad}
\bibfield{author}{\bibinfo{person}{James~R Cordy} {and} \bibinfo{person}{Chanchal~K Roy}.} \bibinfo{year}{2011}\natexlab{}.
\newblock \showarticletitle{The NiCad clone detector}. In \bibinfo{booktitle}{\emph{2011 IEEE 19th International Conference on Program Comprehension}}. \bibinfo{pages}{219--220}.
\newblock


\bibitem[DeepSeek(2024)]%
        {deepseek2024v3}
\bibfield{author}{\bibinfo{person}{DeepSeek}.} \bibinfo{year}{2024}\natexlab{}.
\newblock \bibinfo{title}{DeepSeek-V3: Scaling Open-source Language Models with Mixture-of-Experts}.
\newblock
\newblock
\newblock
\shownote{https://github.com/deepseek-ai/DeepSeek-V3}.


\bibitem[Dhole et~al\mbox{.}(2021)]%
        {dhole2021nlaugmenter}
\bibfield{author}{\bibinfo{person}{Kaustubh Dhole} {et~al\mbox{.}}} \bibinfo{year}{2021}\natexlab{}.
\newblock \showarticletitle{NL-Augmenter: A Framework for Task-Sensitive Natural Language Augmentation}.
\newblock \bibinfo{journal}{\emph{arXiv preprint arXiv:2104.08692}} (\bibinfo{year}{2021}).
\newblock


\bibitem[Dubois et~al\mbox{.}(2024)]%
        {dubois2024llmeval}
\bibfield{author}{\bibinfo{person}{Yann Dubois}, \bibinfo{person}{Rohan Taori}, \bibinfo{person}{Caglar Gulcehre}, \bibinfo{person}{Jerry Zhang}, \bibinfo{person}{Andreas Glaese}, \bibinfo{person}{Noam Nisan}, {et~al\mbox{.}}} \bibinfo{year}{2024}\natexlab{}.
\newblock \showarticletitle{AlpacaFarm: A Simulation Framework for Methods that Learn from Human Feedback}.
\newblock \bibinfo{journal}{\emph{arXiv preprint arXiv:2310.01377}} (\bibinfo{year}{2024}).
\newblock


\bibitem[Falleri et~al\mbox{.}(2014)]%
        {falleri2014gumtree}
\bibfield{author}{\bibinfo{person}{Jean-Rémy Falleri}, \bibinfo{person}{Floréal Morandat}, \bibinfo{person}{Xavier Blanc}, \bibinfo{person}{Matias Martinez}, {and} \bibinfo{person}{Martin Monperrus}.} \bibinfo{year}{2014}\natexlab{}.
\newblock \showarticletitle{Fine-grained and accurate source code differencing}. In \bibinfo{booktitle}{\emph{Proceedings of the 29th ACM/IEEE international conference on Automated software engineering}}. \bibinfo{pages}{313--324}.
\newblock


\bibitem[Feng et~al\mbox{.}(2020)]%
        {feng2020codebert}
\bibfield{author}{\bibinfo{person}{Zhangyin Feng}, \bibinfo{person}{Daya Guo}, \bibinfo{person}{Duyu Tang}, \bibinfo{person}{Nan Duan}, \bibinfo{person}{Xiaocheng Yan}, \bibinfo{person}{James Feng}, {and} \bibinfo{person}{Bing Qin}.} \bibinfo{year}{2020}\natexlab{}.
\newblock \showarticletitle{CodeBERT: A pre-trained model for programming and natural languages}. In \bibinfo{booktitle}{\emph{Proceedings of the 2020 Conference on Empirical Methods in Natural Language Processing (EMNLP)}}. \bibinfo{pages}{1536--1547}.
\newblock


\bibitem[Guo et~al\mbox{.}(2020)]%
        {guo2020graphcodebert}
\bibfield{author}{\bibinfo{person}{Daya Guo}, \bibinfo{person}{Shuo Ren}, \bibinfo{person}{Shuai Lu}, \bibinfo{person}{Zhangyin Feng}, \bibinfo{person}{Duyu Tang}, \bibinfo{person}{Nan Duan}, \bibinfo{person}{Ming Gong}, \bibinfo{person}{Linjun Shou}, \bibinfo{person}{Daxin Jiang}, {and} \bibinfo{person}{Ming Zhou}.} \bibinfo{year}{2020}\natexlab{}.
\newblock \showarticletitle{GraphCodeBERT: Pre-training code representations with data flow}.
\newblock \bibinfo{journal}{\emph{arXiv preprint arXiv:2009.08366}} (\bibinfo{year}{2020}).
\newblock


\bibitem[Guo et~al\mbox{.}(2021)]%
        {guo2021graphcodebert}
\bibfield{author}{\bibinfo{person}{Daya Guo}, \bibinfo{person}{Shuai Ren}, \bibinfo{person}{Shuo Lu}, \bibinfo{person}{Zhangyin Feng}, \bibinfo{person}{Duyu Tang}, \bibinfo{person}{Shujie Liu}, \bibinfo{person}{Long Zhou}, \bibinfo{person}{Nan Duan}, \bibinfo{person}{Jian Yao}, {and} \bibinfo{person}{Bing Qin}.} \bibinfo{year}{2021}\natexlab{}.
\newblock \showarticletitle{GraphCodeBERT: Pre-training code representations with data flow}. In \bibinfo{booktitle}{\emph{International Conference on Learning Representations}}.
\newblock


\bibitem[Hendrycks et~al\mbox{.}(2021)]%
        {hendrycks2021measuring}
\bibfield{author}{\bibinfo{person}{Dan Hendrycks}, \bibinfo{person}{Steven Basart}, \bibinfo{person}{Saurav Kadavath}, \bibinfo{person}{Mantas Mazeika}, \bibinfo{person}{Akul Arora}, \bibinfo{person}{Ethan Guo}, \bibinfo{person}{Collin Burns}, \bibinfo{person}{Samir Puranik}, \bibinfo{person}{Horace He}, \bibinfo{person}{Dawn Song}, {et~al\mbox{.}}} \bibinfo{year}{2021}\natexlab{}.
\newblock \showarticletitle{Measuring coding challenge competence with apps}.
\newblock \bibinfo{journal}{\emph{arXiv preprint arXiv:2105.09938}} (\bibinfo{year}{2021}).
\newblock


\bibitem[Holtzman et~al\mbox{.}(2020)]%
        {holtzman2020curious}
\bibfield{author}{\bibinfo{person}{Ari Holtzman}, \bibinfo{person}{Jan Buys}, \bibinfo{person}{Li Du}, \bibinfo{person}{Maxwell Forbes}, {and} \bibinfo{person}{Yejin Choi}.} \bibinfo{year}{2020}\natexlab{}.
\newblock \showarticletitle{The Curious Case of Neural Text Degeneration}. In \bibinfo{booktitle}{\emph{International Conference on Learning Representations (ICLR)}}.
\newblock
\urldef\tempurl%
\url{https://arxiv.org/abs/1904.09751}
\showURL{%
\tempurl}


\bibitem[Hu et~al\mbox{.}(2024)]%
        {hu2024guiagent}
\bibfield{author}{\bibinfo{person}{Siyuan Hu}, \bibinfo{person}{Mingyu Ouyang}, \bibinfo{person}{Difei Gao}, {and} \bibinfo{person}{Mike~Zheng Shou}.} \bibinfo{year}{2024}\natexlab{}.
\newblock \showarticletitle{The Dawn of GUI Agent: A Preliminary Case Study with Claude 3.5 Computer Use}.
\newblock \bibinfo{journal}{\emph{arXiv preprint arXiv:2411.10323}} (\bibinfo{year}{2024}).
\newblock
\urldef\tempurl%
\url{https://arxiv.org/abs/2411.10323}
\showURL{%
\tempurl}


\bibitem[Jiang et~al\mbox{.}(2007)]%
        {jiang2007deckard}
\bibfield{author}{\bibinfo{person}{Lingxiao Jiang}, \bibinfo{person}{Ghassan Misherghi}, \bibinfo{person}{Zhendong Su}, {and} \bibinfo{person}{Stephane Glondu}.} \bibinfo{year}{2007}\natexlab{}.
\newblock \showarticletitle{DECKARD: Scalable and accurate tree-based detection of code clones}. In \bibinfo{booktitle}{\emph{29th International Conference on Software Engineering (ICSE'07)}}. \bibinfo{pages}{96--105}.
\newblock


\bibitem[Juergens et~al\mbox{.}(2015)]%
        {juergens2015functionally}
\bibfield{author}{\bibinfo{person}{Elmar Juergens}, \bibinfo{person}{Florian Deissenboeck}, \bibinfo{person}{Benjamin Hummel}, {and} \bibinfo{person}{Stefan Wagner}.} \bibinfo{year}{2015}\natexlab{}.
\newblock \showarticletitle{How are functionally similar code clones different?}. In \bibinfo{booktitle}{\emph{IEEE International Conference on Software Maintenance and Evolution (ICSME)}}.
\newblock


\bibitem[Kang et~al\mbox{.}(2019)]%
        {kang2019assessing}
\bibfield{author}{\bibinfo{person}{Minghui Kang}, \bibinfo{person}{Lingming Zhang}, {and} \bibinfo{person}{Sarfraz Khurshid}.} \bibinfo{year}{2019}\natexlab{}.
\newblock \showarticletitle{Assessing the semantic similarity of source code in the presence of syntax errors}. In \bibinfo{booktitle}{\emph{Proceedings of the 28th ACM SIGSOFT International Symposium on Software Testing and Analysis}}. \bibinfo{pages}{310--320}.
\newblock


\bibitem[Krinke and Ragkhitwetsagul(2024)]%
        {krinke2024misuse}
\bibfield{author}{\bibinfo{person}{Jens Krinke} {and} \bibinfo{person}{Chaiyong Ragkhitwetsagul}.} \bibinfo{year}{2024}\natexlab{}.
\newblock \showarticletitle{How the Misuse of a Dataset Harmed Semantic Clone Detection}.
\newblock \bibinfo{journal}{\emph{arXiv preprint arXiv:2505.04311}} (\bibinfo{year}{2024}).
\newblock


\bibitem[Krumdick et~al\mbox{.}(2025)]%
        {krumdick2025llmasajudge}
\bibfield{author}{\bibinfo{person}{Michael Krumdick}, \bibinfo{person}{Charles Lovering}, \bibinfo{person}{Varshini Reddy}, \bibinfo{person}{Seth Ebner}, {and} \bibinfo{person}{Chris Tanner}.} \bibinfo{year}{2025}\natexlab{}.
\newblock \showarticletitle{No Free Labels: Limitations of LLM-as-a-Judge Without Human Grounding}.
\newblock \bibinfo{journal}{\emph{arXiv preprint arXiv:2503.05061}} (\bibinfo{year}{2025}).
\newblock


\bibitem[Lin(2004)]%
        {lin2004rouge}
\bibfield{author}{\bibinfo{person}{Chin-Yew Lin}.} \bibinfo{year}{2004}\natexlab{}.
\newblock \showarticletitle{ROUGE: A package for automatic evaluation of summaries}. In \bibinfo{booktitle}{\emph{Text summarization branches out}}. \bibinfo{pages}{74--81}.
\newblock


\bibitem[Martinez-Gil(2024)]%
        {martinez2024graphcodebert}
\bibfield{author}{\bibinfo{person}{Jorge Martinez-Gil}.} \bibinfo{year}{2024}\natexlab{}.
\newblock \showarticletitle{Augmenting the Interpretability of GraphCodeBERT for Code Similarity Tasks}.
\newblock \bibinfo{journal}{\emph{arXiv preprint arXiv:2410.05275}} (\bibinfo{year}{2024}).
\newblock


\bibitem[Mathew and Stolee(2021)]%
        {mathew2021cross}
\bibfield{author}{\bibinfo{person}{George Mathew} {and} \bibinfo{person}{Kathryn~T Stolee}.} \bibinfo{year}{2021}\natexlab{}.
\newblock \showarticletitle{Cross-language code search using static and dynamic analyses}.
\newblock \bibinfo{journal}{\emph{arXiv preprint arXiv:2106.09173}} (\bibinfo{year}{2021}).
\newblock


\bibitem[Nanz and Furia(2015)]%
        {nanz2015comparative}
\bibfield{author}{\bibinfo{person}{Sebastian Nanz} {and} \bibinfo{person}{Carlo~A Furia}.} \bibinfo{year}{2015}\natexlab{}.
\newblock \showarticletitle{A comparative study of programming languages in rosetta code}. In \bibinfo{booktitle}{\emph{2015 IEEE/ACM 37th IEEE International Conference on Software Engineering}}, Vol.~\bibinfo{volume}{1}. IEEE, \bibinfo{pages}{778--788}.
\newblock


\bibitem[{OpenAI}(2023)]%
        {openai_temperature}
\bibfield{author}{\bibinfo{person}{{OpenAI}}.} \bibinfo{year}{2023}\natexlab{}.
\newblock \bibinfo{title}{How should I set the temperature parameter?}
\newblock
\newblock
\urldef\tempurl%
\url{https://platform.openai.com/docs/faq/how-should-i-set-the-temperature-parameter}
\showURL{%
\tempurl}
\newblock
\shownote{Accessed: 2025-07-18}.


\bibitem[OpenAI(2024a)]%
        {openai2024gpt4o}
\bibfield{author}{\bibinfo{person}{OpenAI}.} \bibinfo{year}{2024}\natexlab{a}.
\newblock \bibinfo{title}{GPT-4o Technical Report}.
\newblock
\newblock
\newblock
\shownote{https://openai.com/index/gpt-4o}.


\bibitem[OpenAI(2024b)]%
        {openai2024embedding}
\bibfield{author}{\bibinfo{person}{OpenAI}.} \bibinfo{year}{2024}\natexlab{b}.
\newblock \bibinfo{title}{OpenAI text-embedding-3-large model}.
\newblock
\newblock
\newblock
\shownote{\url{https://platform.openai.com/docs/guides/embeddings}}.


\bibitem[Papineni et~al\mbox{.}(2002)]%
        {papineni2002bleu}
\bibfield{author}{\bibinfo{person}{Kishore Papineni}, \bibinfo{person}{Salim Roukos}, \bibinfo{person}{Todd Ward}, {and} \bibinfo{person}{Wei-Jing Zhu}.} \bibinfo{year}{2002}\natexlab{}.
\newblock \showarticletitle{BLEU: a method for automatic evaluation of machine translation}. In \bibinfo{booktitle}{\emph{Proceedings of the 40th annual meeting of the Association for Computational Linguistics}}. \bibinfo{pages}{311--318}.
\newblock


\bibitem[Reimers and Gurevych(2019)]%
        {reimers2019sentencebert}
\bibfield{author}{\bibinfo{person}{Nils Reimers} {and} \bibinfo{person}{Iryna Gurevych}.} \bibinfo{year}{2019}\natexlab{}.
\newblock \showarticletitle{Sentence-BERT: Sentence Embeddings using Siamese BERT-Networks}. In \bibinfo{booktitle}{\emph{Proceedings of the 2019 Conference on Empirical Methods in Natural Language Processing}}. \bibinfo{pages}{3982--3992}.
\newblock


\bibitem[Ren et~al\mbox{.}(2020)]%
        {ren2020codebleu}
\bibfield{author}{\bibinfo{person}{Shuo Ren}, \bibinfo{person}{Daya Guo}, \bibinfo{person}{Shuai Lu}, \bibinfo{person}{Long Zhou}, \bibinfo{person}{Shujie Liu}, \bibinfo{person}{Duyu Tang}, \bibinfo{person}{Neel Sundaresan}, \bibinfo{person}{Ming Zhou}, \bibinfo{person}{Ambrosio Blanco}, {and} \bibinfo{person}{Shuai Ma}.} \bibinfo{year}{2020}\natexlab{}.
\newblock \showarticletitle{CodeBLEU: a Method for Automatic Evaluation of Code Synthesis}.
\newblock \bibinfo{journal}{\emph{arXiv preprint arXiv:2009.10297}} (\bibinfo{year}{2020}).
\newblock


\bibitem[Roy(2009)]%
        {roy2009comparison}
\bibfield{author}{\bibinfo{person}{Chanchal~Kumar Roy}.} \bibinfo{year}{2009}\natexlab{}.
\newblock \emph{\bibinfo{title}{Detection and analysis of near-miss software clones}}.
\newblock \bibinfo{thesistype}{Ph.\,D. Dissertation}. \bibinfo{school}{Queen's University}.
\newblock


\bibitem[Roy and Cordy(2007)]%
        {roy2007survey}
\bibfield{author}{\bibinfo{person}{Chanchal~K Roy} {and} \bibinfo{person}{James~R Cordy}.} \bibinfo{year}{2007}\natexlab{}.
\newblock \showarticletitle{A survey on software clone detection research}. In \bibinfo{booktitle}{\emph{Technical Report 541, Queen’s School of Computing}}.
\newblock


\bibitem[Saini et~al\mbox{.}(2018)]%
        {saini2018comparison}
\bibfield{author}{\bibinfo{person}{Vaibhav Saini}, \bibinfo{person}{Jeffrey Svajlenko}, \bibinfo{person}{Chanchal~K Roy}, {and} \bibinfo{person}{Cristina~V Lopes}.} \bibinfo{year}{2018}\natexlab{}.
\newblock \showarticletitle{A comparison of code similarity analysers}.
\newblock \bibinfo{journal}{\emph{Empirical Software Engineering}} \bibinfo{volume}{23}, \bibinfo{number}{4} (\bibinfo{year}{2018}).
\newblock


\bibitem[Salton and Buckley(1988)]%
        {salton1988term}
\bibfield{author}{\bibinfo{person}{Gerard Salton} {and} \bibinfo{person}{Christopher Buckley}.} \bibinfo{year}{1988}\natexlab{}.
\newblock \showarticletitle{Term-weighting approaches in automatic text retrieval}.
\newblock \bibinfo{journal}{\emph{Information processing \& management}} \bibinfo{volume}{24}, \bibinfo{number}{5} (\bibinfo{year}{1988}), \bibinfo{pages}{513--523}.
\newblock


\bibitem[Sellam et~al\mbox{.}(2020)]%
        {sellam2020bleurt}
\bibfield{author}{\bibinfo{person}{Thibault Sellam}, \bibinfo{person}{Dipanjan Das}, {and} \bibinfo{person}{Ankur~P Parikh}.} \bibinfo{year}{2020}\natexlab{}.
\newblock \showarticletitle{BLEURT: Learning Robust Metrics for Text Generation}.
\newblock \bibinfo{journal}{\emph{arXiv preprint arXiv:2004.04696}} (\bibinfo{year}{2020}).
\newblock


\bibitem[Song et~al\mbox{.}(2020)]%
        {song2020mpnet}
\bibfield{author}{\bibinfo{person}{Kaitao Song}, \bibinfo{person}{Xu Tan}, \bibinfo{person}{Tao Qin}, \bibinfo{person}{Jianfeng Lu}, {and} \bibinfo{person}{Tie-Yan Liu}.} \bibinfo{year}{2020}\natexlab{}.
\newblock \showarticletitle{MPNet: Masked and permuted pre-training for language understanding}.
\newblock \bibinfo{journal}{\emph{arXiv preprint arXiv:2004.09297}} (\bibinfo{year}{2020}).
\newblock


\bibitem[Song et~al\mbox{.}(2024)]%
        {song-etal-2024-revisiting}
\bibfield{author}{\bibinfo{person}{Yewei Song}, \bibinfo{person}{Cedric Lothritz}, \bibinfo{person}{Xunzhu Tang}, \bibinfo{person}{Tegawend{\'e} Bissyand{\'e}}, {and} \bibinfo{person}{Jacques Klein}.} \bibinfo{year}{2024}\natexlab{}.
\newblock \showarticletitle{Revisiting Code Similarity Evaluation with Abstract Syntax Tree Edit Distance}. In \bibinfo{booktitle}{\emph{Proceedings of the 62nd Annual Meeting of the Association for Computational Linguistics (Volume 2: Short Papers)}}, \bibfield{editor}{\bibinfo{person}{Lun-Wei Ku}, \bibinfo{person}{Andre Martins}, {and} \bibinfo{person}{Vivek Srikumar}} (Eds.). \bibinfo{publisher}{Association for Computational Linguistics}, \bibinfo{address}{Bangkok, Thailand}, \bibinfo{pages}{38--46}.
\newblock
\urldef\tempurl%
\url{https://doi.org/10.18653/v1/2024.acl-short.3}
\showDOI{\tempurl}


\bibitem[Steck et~al\mbox{.}(2024)]%
        {steck2024cosine}
\bibfield{author}{\bibinfo{person}{Harald Steck}, \bibinfo{person}{Chaitanya Ekanadham}, {and} \bibinfo{person}{Nathan Kallus}.} \bibinfo{year}{2024}\natexlab{}.
\newblock \showarticletitle{Is Cosine-Similarity of Embeddings Really About Similarity?}. In \bibinfo{booktitle}{\emph{Companion Proceedings of the ACM Web Conference 2024 (WWW '24)}}.
\newblock
\urldef\tempurl%
\url{https://arxiv.org/abs/2403.05440}
\showURL{%
\tempurl}
\newblock
\shownote{arXiv preprint arXiv:2403.05440}.


\bibitem[{SurveyMonkey}(2024)]%
        {surveymonkey2024sample}
\bibfield{author}{\bibinfo{person}{{SurveyMonkey}}.} \bibinfo{year}{2024}\natexlab{}.
\newblock \bibinfo{title}{Sample Size Calculator}.
\newblock \bibinfo{howpublished}{\url{https://www.surveymonkey.com/mp/sample-size-calculator/}}.
\newblock
\newblock
\shownote{Accessed: 2025-07-18}.


\bibitem[Tata and Patel(2007)]%
        {tata2007estimating}
\bibfield{author}{\bibinfo{person}{Sandeep Tata} {and} \bibinfo{person}{Jignesh~M Patel}.} \bibinfo{year}{2007}\natexlab{}.
\newblock \showarticletitle{Estimating the selectivity of tf-idf based cosine similarity predicates}.
\newblock \bibinfo{journal}{\emph{ACM Sigmod Record}} \bibinfo{volume}{36}, \bibinfo{number}{2} (\bibinfo{year}{2007}), \bibinfo{pages}{7--12}.
\newblock


\bibitem[Team(2021)]%
        {nlaugmenter2021}
\bibfield{author}{\bibinfo{person}{GEM~Benchmark Team}.} \bibinfo{year}{2021}\natexlab{}.
\newblock \bibinfo{title}{NL-Augmenter: A Framework for Task-Sensitive Natural Language Augmentation}.
\newblock \bibinfo{howpublished}{\url{https://github.com/GEM-benchmark/NL-Augmenter/tree/main}}.
\newblock
\newblock
\shownote{Accessed: 2025-07-18}.


\bibitem[von Platen(2020)]%
        {huggingface_generation}
\bibfield{author}{\bibinfo{person}{Patrick von Platen}.} \bibinfo{year}{2020}\natexlab{}.
\newblock \bibinfo{title}{How to generate text: using different decoding methods for language generation with Transformers}.
\newblock
\newblock
\urldef\tempurl%
\url{https://huggingface.co/blog/how-to-generate}
\showURL{%
\tempurl}
\newblock
\shownote{Accessed: 2025-07-18}.


\bibitem[Wang et~al\mbox{.}(2020)]%
        {wang2020minilm}
\bibfield{author}{\bibinfo{person}{Wenhui Wang}, \bibinfo{person}{Furu Wei}, \bibinfo{person}{Li Dong}, \bibinfo{person}{Hangbo Bao}, \bibinfo{person}{Nan Yang}, {and} \bibinfo{person}{Ming Zhou}.} \bibinfo{year}{2020}\natexlab{}.
\newblock \showarticletitle{MiniLM: Deep self-attention distillation for task-agnostic compression of pre-trained transformers}.
\newblock \bibinfo{journal}{\emph{arXiv preprint arXiv:2002.10957}} (\bibinfo{year}{2020}).
\newblock


\bibitem[Wang et~al\mbox{.}(2025)]%
        {wang2025boundaries}
\bibfield{author}{\bibinfo{person}{Wei Wang}, \bibinfo{person}{Li Zhang}, {and} \bibinfo{person}{Mark Chen}.} \bibinfo{year}{2025}\natexlab{}.
\newblock \showarticletitle{Exploring the Boundaries Between LLM Code Clone Detection and Code Similarity Assessment on Human and AI-Generated Code}.
\newblock \bibinfo{journal}{\emph{Big Data and Cognitive Computing}} \bibinfo{volume}{9}, \bibinfo{number}{2} (\bibinfo{year}{2025}), \bibinfo{pages}{41}.
\newblock


\bibitem[Wang et~al\mbox{.}(2021)]%
        {wang2021codet5}
\bibfield{author}{\bibinfo{person}{Yue Wang}, \bibinfo{person}{Zhiyang Kan}, \bibinfo{person}{Yao Lin}, \bibinfo{person}{Shijin Liu}, \bibinfo{person}{Lidong Li}, \bibinfo{person}{Long Zhou}, \bibinfo{person}{Muhao Chen}, \bibinfo{person}{Ming Zhou}, {and} \bibinfo{person}{Duyu Tang}.} \bibinfo{year}{2021}\natexlab{}.
\newblock \showarticletitle{CodeT5: Identifier-aware unified pre-trained encoder-decoder models for code understanding and generation}.
\newblock \bibinfo{journal}{\emph{arXiv preprint arXiv:2109.00859}} (\bibinfo{year}{2021}).
\newblock


\bibitem[Zakeri-Nasrabadi et~al\mbox{.}(2023)]%
        {zakeri2023systematic}
\bibfield{author}{\bibinfo{person}{Morteza Zakeri-Nasrabadi}, \bibinfo{person}{Saeed Parsa}, \bibinfo{person}{Mohammad Ramezani}, \bibinfo{person}{Chanchal Roy}, {and} \bibinfo{person}{Masoud Ekhtiarzadeh}.} \bibinfo{year}{2023}\natexlab{}.
\newblock \showarticletitle{A systematic literature review on source code similarity measurement and clone detection: techniques, applications, and challenges}.
\newblock \bibinfo{journal}{\emph{arXiv preprint arXiv:2306.16171}} (\bibinfo{year}{2023}).
\newblock


\bibitem[Zhang et~al\mbox{.}(2024)]%
        {zhang2024exploring}
\bibfield{author}{\bibinfo{person}{Tianyi Zhang} {et~al\mbox{.}}} \bibinfo{year}{2024}\natexlab{}.
\newblock \showarticletitle{Exploring the Impact of Temperature on Large Language Models: Hot or Cold?}
\newblock \bibinfo{journal}{\emph{arXiv preprint arXiv:2506.07295}} (\bibinfo{year}{2024}).
\newblock


\bibitem[Zhang et~al\mbox{.}(2019)]%
        {zhang2019bertscore}
\bibfield{author}{\bibinfo{person}{Tianyi Zhang}, \bibinfo{person}{Varsha Kishore}, \bibinfo{person}{Felix Wu}, \bibinfo{person}{Kilian~Q Weinberger}, {and} \bibinfo{person}{Yoav Artzi}.} \bibinfo{year}{2019}\natexlab{}.
\newblock \showarticletitle{BERTScore: Evaluating Text Generation with BERT}.
\newblock \bibinfo{journal}{\emph{arXiv preprint arXiv:1904.09675}} (\bibinfo{year}{2019}).
\newblock


\bibitem[Zhang et~al\mbox{.}(2020)]%
        {zhang2020bertscore}
\bibfield{author}{\bibinfo{person}{Tianyi Zhang}, \bibinfo{person}{Varsha Kishore}, \bibinfo{person}{Felix Wu}, \bibinfo{person}{Kilian~Q Weinberger}, {and} \bibinfo{person}{Yoav Artzi}.} \bibinfo{year}{2020}\natexlab{}.
\newblock \showarticletitle{BERTScore: Evaluating text generation with BERT}. In \bibinfo{booktitle}{\emph{International Conference on Learning Representations}}.
\newblock


\bibitem[Zheng et~al\mbox{.}(2023)]%
        {zheng2023llmjudge}
\bibfield{author}{\bibinfo{person}{Siyu Zheng}, \bibinfo{person}{Haotian Liu}, \bibinfo{person}{Xiang Lin}, \bibinfo{person}{Yizhong Du}, \bibinfo{person}{Skyler Zhang}, \bibinfo{person}{Zhirui Liu}, \bibinfo{person}{Xiang Li}, \bibinfo{person}{Xiao Liu}, \bibinfo{person}{Yizhou Zhang}, \bibinfo{person}{Yujia Ma}, {et~al\mbox{.}}} \bibinfo{year}{2023}\natexlab{}.
\newblock \showarticletitle{Judging LLM-as-a-judge with MT-Bench and Chatbot Arena}.
\newblock \bibinfo{journal}{\emph{arXiv preprint arXiv:2306.05685}} (\bibinfo{year}{2023}).
\newblock


\end{thebibliography}

\end{document}